\documentclass[a4paper,fleqn]{cas-sc}

\PassOptionsToPackage{colorlinks,hidelinks}{hyperref}
\usepackage{hyperref}
\usepackage{cleveref}

\usepackage[authoryear]{natbib}
\usepackage{graphicx}
\usepackage{amsmath}
\usepackage{caption}
\usepackage{subcaption}
\usepackage{tocloft}
\usepackage{multirow} 
\usepackage{pdflscape} % Include this package for landscape pages

\def\tsc#1{\csdef{#1}{\textsc{\lowercase{#1}}\xspace}}
\tsc{WGM}
\tsc{QE}
\begin{document}
\let\WriteBookmarks\relax
\def\floatpagepagefraction{1}
\def\textpagefraction{.001}

\shorttitle{}    

\shortauthors{Tang, Wang, and Delle Monache}  

\title [mode = title]{Strategizing Equitable Transit Evacuations: A Data-Driven Reinforcement Learning Approach}  

% Parsimonious Resilience Dynamics Analysis Beyond V-shapes: Coupled Ordinary Differential Equations with Double Quadratic Queue Model

%\journal{Transportation Research - Part C}
% % Title footnote mark
% % eg: \tnotemark[1]

% % Title footnote 1.
% % eg: \tnotetext[1]{Title footnote text}


\author[1,2]{Fang Tang}[orcid=0009-0002-2468-9554]

\ead{fangt@asu.edu}

% \ead[url]{https://orcid.org/0009-0002-2468-9554}

\credit{}

\affiliation[1]{organization={School of Sustainable Engineering and the Built Environment, Arizona State University},
            % addressline={660 S. College Avenue}, 
            city={Tempe},
%          citysep={}, % Uncomment if no comma needed between city and postcode
            % postcode={85287}, 
            state={AZ},
            country={USA}}

\author[2]{Han Wang}[orcid=0000-0003-1561-7532]

\ead{hanw@berkeley.edu}

% \ead[url]{https://orcid.org/0009-0002-2468-9554}

\credit{}

\affiliation[2]{organization={Department of Civil and Environmental Engineering,         University of California, Berkeley},
            % addressline={660 S. College Avenue}, 
            city={Berkeley},
%          citysep={}, % Uncomment if no comma needed between city and postcode
            % postcode={85287}, 
            state={CA},
            country={USA}}
            
\author[2]{Maria Laura {Delle Monache}}[orcid=0000-0003-1355-8153]

\cormark[1]

% Email id of the second author
\ead{mldellemonache@berkeley.edu}

% \ead[url]{https://orcid.org/0000-0002-9963-5369}

\credit{}

% % Address/affiliation
% \affiliation[2]{organization={},
% %          citysep={}, % Uncomment if no comma needed between city and postcode

\cortext[1]{Corresponding author}

% % Footnote text


\begin{abstract}
As natural disasters become increasingly frequent, the need for efficient and equitable evacuation planning has become more critical. This paper proposes a data-driven, reinforcement learning (RL)-based framework to optimize bus-based evacuations with an emphasis on improving both efficiency and equity. We model the evacuation problem as a Markov Decision Process (MDP) solved by RL, using real-time transit data from General Transit Feed Specification (GTFS) and transportation networks extracted from OpenStreetMap (OSM). The RL agent dynamically reroutes buses from their scheduled location to minimize total passengers’ evacuation time while prioritizing equity priority communities. Simulations on the San Francisco Bay Area transportation network indicate that the proposed framework achieves significant improvements in both evacuation efficiency and equitable service distribution compared to traditional rule-based and random strategies. These results highlight the potential of RL to enhance system performance and urban resilience during emergency evacuations, offering a scalable solution for real-world applications in intelligent transportation systems.
\end{abstract}

\begin{highlights}
\item Proposes a Reinforcement Learning-based bus evacuation planning.
\item Integrates multiple sources of data from OpenStreetMap and General Transit Feed Specification.
\item Incorporates equity in evacuation strategy.
\item Demonstrates scalability and applicability on real-world urban large-scale networks.
\end{highlights}

% Keywords

\maketitle

% Main text
\section{Introduction}\label{Introduction}
The increasing frequency of natural disasters, such as wildfires, flooding, heat waves, hurricanes, and earthquakes, poses significant risks to residents' finances, society, and safety. Efficient and real-time evacuation planning strategies are essential not only to save costs but also to protect human lives. Buses, with their flexibility and large capacity, offer a cost-efficient solution for evacuations. \cite{khalili2024systematic} conducted a comprehensive review identifying key challenges and research gaps in transit-based evacuation planning. This review highlights the critical role of integrating public transportation when developing evacuation strategies, in particular when enhancing efficiency and safety. Evacuation plans based on transit are crucial for individuals without personal vehicles and for vehicle owners during disasters when cars may be inaccessible. The review emphasizes the need to consider spatial and temporal dynamics, social and demographic factors, and multi-stakeholder coordination. Despite advancements, challenges still persist as for example data accuracy, model validation, and stakeholder coordination, necessitating further research and interdisciplinary collaboration to improve transit-based evacuation strategies.

Previous studies on bus evacuation problems, such as those by \cite{HamacherTjandra2001}, \cite{altay2006or}, \cite{yusoff2008optimization}, and \cite{goerigk2013branch}, primarily focus on optimizing evacuation time. \cite{chen2008agent} considered buses as agents in optimizing bus-based evacuation planning. \cite{sayyady2010optimizing} addressed the bus evacuation problem considering traffic flow dynamics to model real-world complexity. \cite{goerigk2014robust} studied uncertain bus evacuation problems, modeling the uncertainty in the number of evacuees. \cite{goerigk2013branch} proposed a branch-and-bound algorithm to optimize total evacuation time for the problem with a single bus depot.

\cite{huang2016leaves} highlighted the importance of decision variables in emergency evacuation planning, focusing on the allocation of evacuees and resources. In transit-based evacuation planning, decision variables often include vehicle routing, scheduling, and allocation, as well as the location and allocation of shelters and pick-up points. Some studies also incorporate relief supply and distribution planning. Objective functions typically aim to minimize evacuation time, transportation costs \citep{bish2011planning}, and evacuation risks while maximizing the number of evacuees transported safely. Other objectives include minimizing the network clearance time \citep{goerigk2013branch}.

% The complexity of real-world disaster scenarios often necessitates multi-objective models that address various constraints and uncertainties, such as shelter capacities, vehicle availability, and network conditions.

Optimization approaches in transit-based evacuation planning include dynamic network methods, time-dependent flow models, heuristic algorithms, and stochastic programming frameworks. For example, \cite{bish2011planning} developed mixed integer programming models and a local search heuristic for efficient evacuation planning, aiming to optimize resource allocation, route planning, and scheduling under various constraints and uncertainties. \cite{khalili2024systematic} underscored the effectiveness of integrating optimization with traffic simulation and metaheuristic methods to enhance evacuation efficiency. Despite advancements, challenges remain in model validation, data accuracy, and stakeholder coordination, necessitating continued research and development in this field. \cite{lu2005capacity} proposed capacity-constrained routing algorithms, and \cite{cova2003network} introduced lane-based evacuation routing. Recent studies have shown that machine learning (ML) can optimize complex decision-making processes in dynamic environments. \cite{wang2022urban} introduced equity-focused recovery strategies using reinforcement learning (RL) to enhance resilience to urban networks, demonstrating the potential of RL to balance efficiency and equity in transportation networks.

Various studies were conducted on bus evacuation planning, with focus on optimization models \citep{liu2006two}, simulation tools for emergency evacuation \citep{chen2009modeling} and multiobjective optimization \citep{abdelgawad2010multiobjective} or on assessing multimodal evacuation protocols for special needs populations \citep{kaiser2012emergency}. Vulnerable communities are often overlooked in public transportation evacuations, suggesting that governmental agencies should address social equity as noted in \cite{litman2006lessons}, \cite{renne2011carless}, and \cite{wu2012logistics}. Equity-based evacuation planning ensures that equity priority communities have equal access to evacuation resources and services. Effective disaster management should ensure equal treatment for all communities. In addition, most research relies on static data and predefined scenarios, which may not adapt well to real-time disaster conditions.

To evacuate populations from multiple communities to designated shelters while ensuring equitable service, this paper proposes an equity-emphasized bus evacuation problem. We utilize a data-driven approach based on the Markov decision framework and RL to reroute or reschedule buses. Real-time bus locations are obtained from the \cite{gtfs_data_transitfeeds} during a disaster. Buses are rerouted through the transportation network extracted from \cite{OpenStreetMap} to evacuate populations. An integrated objective combining an equity evaluation index and passenger travel time serves as a reward function for the RL agent to determine the evacuation priority of affected communities and guide bus routing.

The paper is organized as follows. Section \ref{sec:methodology} describes the network modeling, the bus evacuation problem with equity considerations, the Markov Chain framework, and the adopted equity index. Section \ref{sec:problem statement} uses a six-node network to illustrate the problem and feasible solutions. Section \ref{sec:simulation} introduces the datasets used and the reinforcement learning environment built for San Francisco scenarios. Finally, Section \ref{sec:results} presents the results of the different experimental scenarios.

\section{Methodology}\label{sec:methodology}
In response to the limitations of existing evacuation frameworks, particularly in addressing service equity and operational timeliness, we propose a data-driven, RL-based Markov decision framework. This approach integrates real-time transit data and road network to dynamically adjust bus routing and scheduling during emergencies, ensuring both operational efficiency and equitable service distribution. 

\subsection{Network modeling}
The transportation network is modeled as a directed graph \( G = (V, E, B, I) \), where \( V \) represents the set of nodes, including origins, shelters, and normal transit nodes, and \( E \) represents the set of directed links between these nodes. Each node \( v \in V \) is characterized by its coordinates, type (origin, shelter, or transit), demand (number of evacuees to be transferred) $\lambda(v)$, and remaining capacity of shelter $\mu(v)$. At origins, $\mu(v)$ is always 0, and at shelters, $\lambda(v)$ is always 0. Each link \( e \in E \) is characterized by its travel time $t(e)$, direction, and connectivity between nodes. Each bus $b$ for $b\in B$ is featured by its capacity $c(b)$, number of evacuees on board $p(b)$, and the destination of current route $d(b)$. The nodes and links in the transportation network are mapped to the census tracts $i\in I$ (refer to Figure \ref{fig:census_tracts}), where link \( e_i \in E(i) \) and node \( v_i \in V(i) \).

\subsection{Markov Decision Process for evacuation problem}
\begin{figure}[thpb]
    \centering
    \includegraphics[scale=0.25]{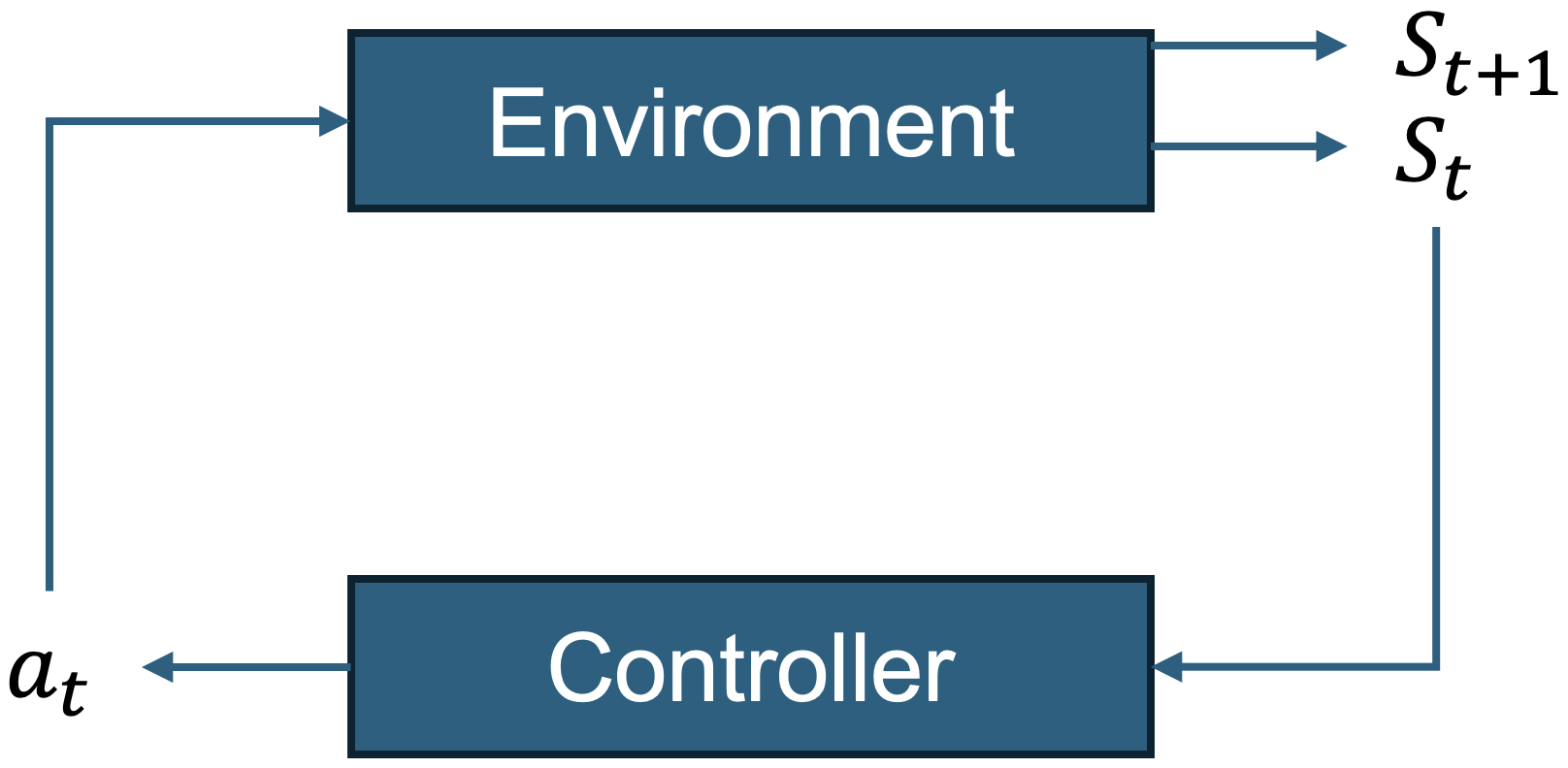}
    \caption{The framework of the proposed methodology using reinforcement learning. The RL agent (controller) interacts with the environment, which represents the transportation network. At each time step \( t \), the controller takes an action \( a_t \) based on the current state \( S_t \). The environment then transitions to a new state \( S_{t+1} \), providing feedback to the controller. }
    \label{fig_markov_chain}
\end{figure}

As shown in Figure \ref{fig_markov_chain}, the evacuation problem is formulated as a Markov Decision Process defined by the tuple \( (\mathcal{S}, \mathcal{A}, \mathcal{R}, \gamma) \), where:

\begin{itemize}
    \item \textbf{State Space} \( \mathcal{S} \): The state \( S_t \) at time \( t \) includes the locations of all buses represented by $\{s_b, s_v\}$. 
    $s_b = \{c(b), p(b), d(b) | b \in B\}$ defines the status of all buses, and $s_v = \{ \lambda(v), \mu(v) | v \in V\}$ indicates the number of evacuees at each origin, and the remaining capacity of each shelter at time $t$. 
    \item \textbf{Action Space} \( \mathcal{A} \): The action $a_t(b) \in V$ taken by the agent includes decisions to reroute buses, specifying which bus should move to which node as the destination for the next time step.
    \item \textbf{Reward Function} \( \mathcal{R} \): The reward function \( R(S_t, a_t) \) combines the travel time and an equity evaluation index, designed to minimize the total evacuee waiting and travel time while penalizing inequities. The following section provides detailed information of the reward function.
    \item \textbf{Discount Factor} \( \gamma \): The discount factor \( \gamma \in [0,1) \) determines the present value of future rewards.
\end{itemize}

\subsection{Equity emphasized rewards}
The overall reward \( R \) minimizes the total evacuee waiting and travel time with an inequity penalty:
\begin{align}
    R = T + J
\end{align}
\begin{align}
    T = (\sum_{b\in B}p(b) + \sum_{i \in I}\sum_{v_i\in V(i)} \lambda(v_i)) \Delta t 
\end{align}
with $T$, the total travel and waiting time for evacuees with a fixed time interval $\Delta t$. This function facilitates an analysis at the node level, allowing researchers to incorporate specific demand data per node if available. In the absence of such data, it is permissible to allocate the average demand from zone $i$ to each corresponding node $v_i$. $J$ is the inequity penalty. Inspired by the penalty measure in \cite{wang2022urban}, the inequity penalty is calculated as follows: 
\begin{equation}
J=|r_{pb}(\lambda_{i},X_{i}^{e})| \cdot T,i\in I
\label{eqn_J}
\end{equation}
where $r_{pb}(\lambda_i,X_{i}^{e})$ is the point-biserial coefficient between the evacuee demand $\lambda_{i}=\sum_{v_i\in V(i)} \lambda(v_i)$ of zone $i$, and the binary variable $X_{i}^{e}$ that indicates if the zone $i$ is a \textit{equity priority community (EPC)} \citep{epc} or not. EPC refers to tracts with a high proportion of equity-priority communities. Various criteria can be applied to identify an EPC, which is then represented by the binary variable \( X_{i}^{e} \in \{0,1\} \). The point-biserial correlation coefficient is then calculated as:

\begin{equation}
r_{pb}(\lambda_{i},X_{i}^{e})=
\frac {\overline{\lambda}_{e}-\overline{\lambda}_{ne}}{s_{n}} \sqrt{\frac {n_{e}n_{ne}}{n^{2}}} \qquad i \in S.
\label{eqn_r}
\end{equation}

In equation \eqref{eqn_r}, zones are divided into two groups: EPC zones, denoted by subscript \( e \), and non-EPC zones, denoted by subscript \( ne \). Here, \( \overline{\lambda}_{e} \) and \( \overline{\lambda}_{ne} \) represent the mean evacuee demand in the EPC and non-EPC groups, respectively, while \( n_{e} \) and \( n_{ne} \) denote the number of zones in each group, and \( n \) is the total number of zones. The standard deviation of evacuee demand across all zones is given by \( s_{n} \). This point-biserial correlation coefficient quantifies the linear relationship between evacuee demand and EPC designation, indicating whether there are significant differences in evacuation demand between EPC zones and other areas.

% EPC refers to the tract in which there is a significant proportion of equity priority communities. To determine an EPC different criteria can be used. Once the EPC is determined and noted by the variable $X_{i}^{e}\in\{0,1\}$, the point-biserial coefficient is obtained by:
% \begin{equation}
% r_{pb}(\lambda_{i},X_{i}^{e})=
% {\frac  {\overline{\lambda}_{e}-\overline{\lambda}_{ne}}{s_{n}}}{\sqrt{{\frac  {n_{e}n_{ne}}{n^{2}}}}} \qquad i\in S .
% \label{eqn_r}
% \end{equation}

% In \eqref{eqn_r},  the zones are divided into two groups to distinguish an EPC with subscript \textit{e} with one that is not with the subscript \textit{ne}. $\overline{\lambda}_{e}$ and $\overline{\lambda}_{ne}$ are the mean values of the evacuee demand within the two groups. $n_{e}$ and $n_{ne}$ are the numbers of zones in each group and $n$ is the total number of zones. $s_{n}$ is the standard deviation of the evacuee demand on the whole sample. The point-biserial coefficient is the measure of the linear relationship between the value of evacuee demand and the EPC. It indicates whether there are significant differences in evacuation demand between these EPCs and other communities.

The reward function incentivizes the RL agent to prioritize actions that balance efficiency and equity, ensuring that the evacuation process is fair and effective.

\subsection{Reinforcement Learning controller}
\label{subsec:algorithm}

To address the bus evacuation problem, the Proximal Policy Optimization (PPO) algorithm is employed to train the RL agent due to its stability and efficiency within policy gradient methods. This section delineates the MDP formulation, the PPO algorithm, and the training process in detail.

The PPO algorithm is a policy gradient method that directly optimizes the policy \( \pi_{\theta}(a|s) \). The primary objective is to maximize the expected cumulative reward. The PPO algorithm is characterized by its clipped surrogate objective, which constrains updates to the policy to prevent substantial deviations from the previous policy, thereby maintaining stability.

\subsubsection{Clipped surrogate objective}
The objective function for PPO is formulated as:
\begin{equation}
L^{\text{CLIP}}(\theta) = \mathbb{E}_t \left[ \min \left( r_t(\theta) \hat{A}_t, \text{clip}(r_t(\theta), 1 - \epsilon, 1 + \epsilon) \hat{A}_t \right) \right]
\end{equation}
where:
\begin{itemize}
    \item \( r_t(\theta) = \frac{\pi_{\theta}(a_t|s_t)}{\pi_{\theta_{\text{old}}}(a_t|s_t)} \) is the probability ratio between the new policy \( \pi_{\theta} \) and the old policy \( \pi_{\theta_{\text{old}}} \).
    \item \( \hat{A}_t \) is the advantage estimate at time \( t \), computed to reduce variance in the policy gradient.
    \item \( \epsilon \) is a hyperparameter that determines the clipping range.
\end{itemize}

The clipping term \( \text{clip}(r_t(\theta), 1 - \epsilon, 1 + \epsilon) \) ensures that the ratio \( r_t(\theta) \) does not deviate significantly from 1, thereby preventing large policy updates that could destabilize training.

\subsubsection{Advantage estimation}
The advantage estimate \( \hat{A}_t \) is computed using the Generalized Advantage Estimation (GAE), which provides a method to calculate the advantage function \( \hat{A}_t \) by weighting the temporal difference residuals:
\begin{equation}
\hat{A}_t = \sum_{l=0}^{\infty} (\gamma \lambda)^l \delta_{t+l}
\end{equation}
where:
\begin{itemize}
    \item \( \delta_t = R_t + \gamma V(s_{t+1}) - V(s_t) \) is the temporal difference error, with \( R_t \) representing the reward at time \( t \), \( \gamma \) the discount factor, and \( V(s_t) \) the value function.
    \item \( \lambda \) is the GAE parameter that controls the bias-variance trade-off.
\end{itemize}

The GAE method improves the bias-variance trade-off by adjusting \( \lambda \), thereby providing a more stable and reliable estimate of the advantage function.

\subsection{Training process}
The training process involves iterative interactions with the environment, collection of trajectories, and updates to the policy. The detailed steps are as follows:

\begin{enumerate}
    \item \textbf{Initialize the policy} \( \pi_{\theta} \) and value function \( V_{\phi} \) with random parameters \( \theta \) and \( \phi \).
    \item \textbf{Collect trajectories} by executing the current policy \( \pi_{\theta} \) within the environment. Each trajectory \( \tau \) consists of states \( s_t \), actions \( a_t \), rewards \( R_t \), and next states \( s_{t+1} \).
    \item \textbf{Compute rewards} \( R_t \) and advantage estimates \( \hat{A}_t \) for each time step in the trajectory.
    \item \textbf{Update the policy} by maximizing the clipped surrogate objective \( L^{\text{CLIP}}(\theta) \) using gradient ascent. Simultaneously, update the value function \( V_{\phi} \) by minimizing the value loss \( L^{\text{VF}}(\phi) \):
    \begin{equation}
    L^{\text{VF}}(\phi) = \mathbb{E}_t \left[ (V_{\phi}(s_t) - R_t)^2 \right]
    \end{equation}
    \item \textbf{Repeat} the process until convergence or until a predefined number of iterations is reached.
\end{enumerate}

The overall training objective combines the policy loss \( L^{\text{CLIP}}(\theta) \), the value function loss \( L^{\text{VF}}(\phi) \), and an entropy bonus \( S[\pi_{\theta}] \) to encourage exploration and avoid premature convergence to suboptimal policies:
\begin{equation}
L(\theta, \phi) = L^{\text{CLIP}}(\theta) - c_1 L^{\text{VF}}(\phi) + c_2 S[\pi_{\theta}]
\end{equation}
where \( c_1 \) and \( c_2 \) are coefficients that balance the contributions of the value loss and the entropy bonus.

In the subsequent sections, we will illustrate the equity-emphasized bus evacuation problem using a six-node network and validate the proposed approach through simulations on the San Francisco Bay Area network. The simulations will leverage real-time bus location data from the GTFS and the transportation network extracted from OSM.

\section{Illustrative Examples}
\label{sec:problem statement}
To provide a detail illustration of how the proposed RL framework operates in the context of bus evacuation planning, this section presents an example using a simplified six-node network scenario. This controlled setting is specifically designed to clarify the problem space and illustrate feasible solutions, without the application of the RL controller. The purpose is to provide a foundational understanding of the evacuation dynamics under simplified conditions, without the complexity of larger real-world networks, which helps in visualizing how the model handles various aspects of the problem, such as route selection and capacity management. This example serves to bridge the conceptual gap to more complex real-world applications detailed in subsequent sections. Section 3.1 introduces how we represent networks. Section 3.2 discusses two feasible solutions without applying the RL framework, we focus instead on the potential impacts of equity. Section 3.3 introduces the application of the RL framework in this six-node example. 

% The RL controller is not applied in this simple demo and we focus instead on the basic principles and objectives of the bus evacuation problem to ease the comprehension of the model's potential impacts before tackling the advanced simulations that incorporate the full capabilities of the RL framework.

\begin{figure}[thpb]
    \centering
    \includegraphics[scale=0.45]{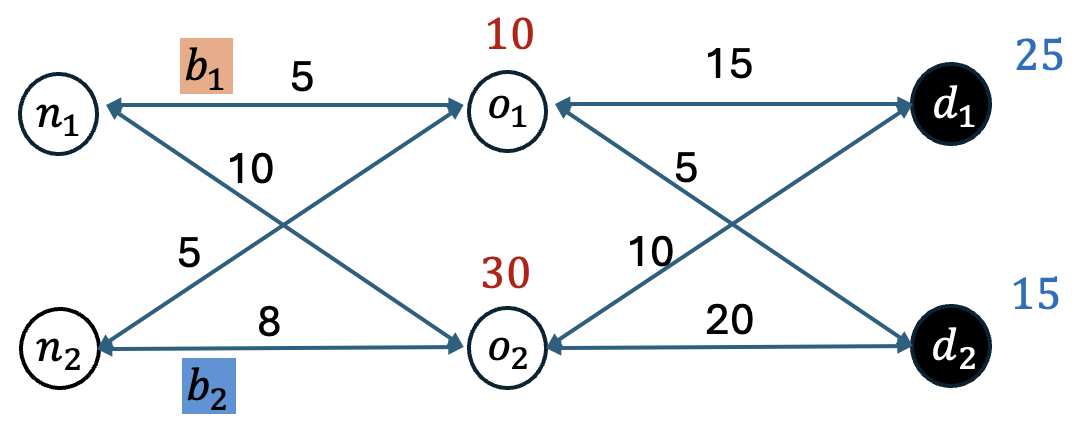}
    \caption{Illustrative example of a six-node network. This figure presents a network comprising six nodes interconnected by directed links, illustrated by the lines with arrows indicating direction. Nodes are denoted as circles, while buses are depicted as square blocks positioned on the links. The numbers on the links represent travel times. Red numerals indicate the demand of evacuees at origin nodes, and blue numerals represent the capacity of shelters.}
    \label{fig_1}
\end{figure}

As shown in Figure \ref{fig_1}, the six-node network with two buses we consider is represented as follows:
\begin{align}
    G_o &= (V_o, E_o, B_o, I_o) \\
    V_o &= \{o_1, o_2, d_1, d_2, n_1, n_2\} \\
    E_o &= \{1, 2, \ldots, 16\} \\
    B_o &= \{b_1, b_2\}\\
    I_o &= \{o_1, o_2\}
\end{align}
This network consists of two endangered communities, denoted as origin nodes \( o_1 \) and \( o_2 \), and two shelters, denoted as destination nodes \( d_1 \) and \( d_2 \). Nodes \( n_1 \) and \( n_2 \) are driving network nodes without capacity or demand. The numbers in set \( E_o \) represent links ID. Two buses, \( b_1 \) and \( b_2 \), are initially located at the link from node \( n_1 \) to origin \( o_1 \) and the link from origin \( o_2 \) to node \( n_2 \) respectively.

In this network, there are 10 and 30 evacuees of origin \( o_1 \) and \( o_2 \) respectively. The capacity for shelters \( d_1 \) and \( d_2 \) are 25 and 15 respectively. The objective is to evacuate all residents of the origin communities to shelters with capacity constraints while ensuring that the evacuation process is equitable for all communities involved.

\subsection{Network specification} \label{subsec: Network Specification}

The nodes in the network are identified by their unique \textit{node\_id}. Table \ref{tab:node_information} provides detailed information for each node, including its x and y coordinates, type, demand (number of people waiting for evacuation), capacity (number of people the shelter can accommodate), and inequity index. The geometry of each node is represented by its x and y coordinates. The \textit{node\_type} specifies whether the node is an origin, shelter, or other types. The nodes designated as origins have a positive demand, while those shelter nodes have a positive capacity. For all other nodes, both demand and capacity are set to 0, indicating that they neither serve as origins nor shelters for evacuees. The \textit{inequity\_index} is used to identify vulnerable communities, where a value of 1 signifies an EPC and 0 otherwise.

As shown in Table \ref{tab:link_information}, each link in the network is a one-way segment characterized by its \textit{link\_id}, \textit{from\_node\_id}, \textit{to\_node\_id}, and \textit{travel\_time}. Each \textit{link\_id} uniquely identifies a specific link. The \textit{from\_node\_id} indicates the starting node of the link, while the \textit{to\_node\_id} represents the end node. The travel time required for buses to traverse each link is recorded in the field \textit{travel\_time}. For example, link 1 starts from node $n_1$ to node $o_1$, requiring 5 minutes of travel time for buses traveling through this link. 

Each bus in the network is identified by its \textit{bus\_id} and key attributes, including its initial location, remaining capacity, and its next destination. Table \ref{tab:bus_information} provides the details of each bus. The initial location is specified by the \textit{link\_id}, \textit{from\_node\_id}, and \textit{to\_node\_id}, along with the \textit{time\_to\_travel} which is the travel time from the initial location to \textit{to\_node\_id}. The capacity indicates the remaining seats available on each bus. The destination define the next destination for bus location at time $t$. 

\begin{table*}[!H]
\caption{Node information}
\label{tab:node_information}
\begin{center}
\begin{tabular}{p{2cm}p{1.5cm}p{1.5cm}p{1.5cm}p{1.5cm}p{1.5cm}p{1.5cm}p{2cm}}
\toprule
\textbf{name} & \textbf{node\_id} & \textbf{x\_coord} & \textbf{y\_coord} & \textbf{node\_type} & \textbf{demand} & \textbf{capacity} & \textbf{inequity\_index}
\\
\hline
node 1 & \(n_1\) & 0 & 1 & node & 0 & 0 & 0 \\
node 2 & \(n_2\) & 0 & 0 & node & 0 & 0 & 0 \\
origin 1 & \(o_1\) & 1 & 1 & origin & 10 & 0 & 1 \\
origin 2 & \(o_2\) & 1 & 0 & origin & 30 & 0 & 0 \\
shelter 1 & \(d_1\) & 2 & 1 & shelter & 0 & 25 & 0 \\
shelter 2 & \(d_2\) & 2 & 0 & shelter & 0 & 15 & 0 \\
\bottomrule
\end{tabular}
\end{center}

\caption{Link information}
\label{tab:link_information}
\begin{center}
\begin{tabular}{p{1.5cm}p{2cm}p{1.5cm}p{1.5cm}}
\toprule
\textbf{link\_id} & \textbf{from\_node\_id} & \textbf{to\_node\_id} & \textbf{travel\_time} \\
\hline
        1 & \(n_1\) & \(o_1\) & 5 \\
        2 & \(n_1\) & \(o_2\) & 10 \\
        3 & \(n_2\) & \(o_1\) & 5 \\
        4 & \(n_3\) & \(o_2\) & 8 \\
        5 & \(o_1\) & \(n_1\) & 5 \\
        6 & \(o_1\) & \(n_2\) & 5 \\
        7 & \(o_1\) & \(d_1\) & 15 \\
        8 & \(o_1\) & \(d_2\) & 5 \\
        9 & \(o_2\) & \(n_1\) & 10 \\
        10 & \(o_2\) & \(n_2\) & 8 \\
        11 & \(o_2\) & \(d_1\) & 10 \\
        12 & \(o_2\) & \(d_2\) & 20 \\
        13 & \(d_1\) & \(o_1\) & 15 \\
        14 & \(d_1\) & \(o_2\) & 10 \\
        15 & \(d_2\) & \(o_1\) & 5 \\
        16 & \(d_2\) & \(o_2\) & 20 \\
\bottomrule
\end{tabular}
\end{center}

\caption{Bus information}
\label{tab:bus_information}
\begin{center}
\begin{tabular}{p{2cm}p{1.5cm}p{2cm}p{2cm}p{2cm}p{2cm}p{2cm}}
\toprule
\textbf{bus\_id} & \textbf{link\_id} & \textbf{from\_node\_id} & \textbf{to\_node\_id} & \textbf{time\_to\_travel} & \textbf{capacity}& \textbf{destination}
 \\
\hline
        \(b_1\) & 1 & \(n_1\) & \(o_1\) & 4 & 20 & \(o_1\) \\
        \(b_2\) & 10 & \(o_2\) & \(n_2\) & 5 & 20 & \(n_2\) \\
\bottomrule
\end{tabular}
\end{center}
\end{table*}

\begin{table*}[!h]
\caption{Feasible solutions for the six-node network}
\label{tab:feasible_solutions}
\begin{center}
\begin{tabular}{p{2.5cm}p{4.5cm}p{2.5cm}p{2.5cm}p{2.5cm}}
\toprule
\textbf{Bus} & \textbf{Trips} &  \textbf{Trip Time} &  \textbf{Waiting Time} & \textbf{Passengers} \\
\hline
\multicolumn{5}{l}{Feasible Solution 1, Total passenger time = $19 \times 10 + 39 \times 15 + 33 \times 15 = 1,270$} \\
\hline
Bus 1 & $(b_1, o_1, d_1)$ & 15 & 4 & 10 \\ 
Bus 1 & $(d_1, o_2, d_1)$ & 10 & 29 & 15 \\ 
Bus 2 & $(b_2, n_2, o_2, d_2)$ & 20 & 13 & 15 \\ 
\hline
\multicolumn{5}{l}{Feasible Solution 2, Total passenger time = $19 \times 10 + 49 \times 15 + 23 \times 15 = 1,270$} \\
\hline
Bus 1 & $(b_1, o_1, d_1)$ & 15 & 4 & 10 \\ 
Bus 1 & $(d_1, o_2, d_2)$ & 20 & 29 & 15 \\ 
Bus 2 & $(b_2, n_2, o_2, d_1)$ & 10 & 13 & 15 \\ 
\bottomrule
\end{tabular}
\end{center}
\end{table*}

\subsection{Feasible solutions}\label{Feasible solutions}
To illustrate potential strategies for the bus evacuation problem with an inequitable origin, \(o_1\), in the six-node network example, we consider two feasible evacuation strategies, as outlined in Table \ref{tab:feasible_solutions}. The total passenger time is computed as the sum of both the trip time and waiting time for all passengers. The trip time refers to the duration of transportation from the origin to the destination, while the waiting time is measured from the moment the hazard occurs until the passenger is picked up.

\begin{figure*}[!ht]
    \centering
    \subfloat[Feasible solution 1]
    {\includegraphics[width=0.45\columnwidth]{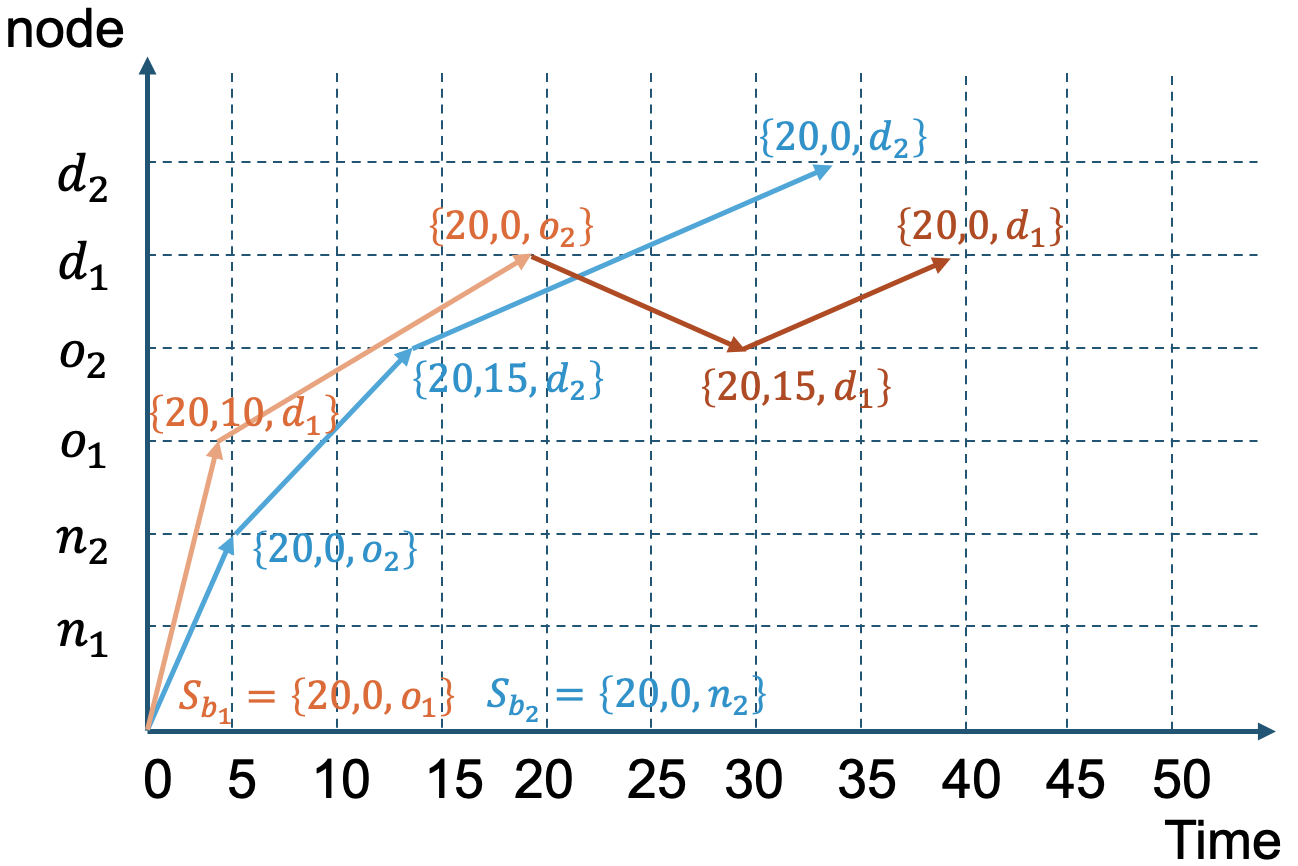}} \hspace{5pt}
    \subfloat[Feasible solution 2]
    {\includegraphics[width=0.45\columnwidth]{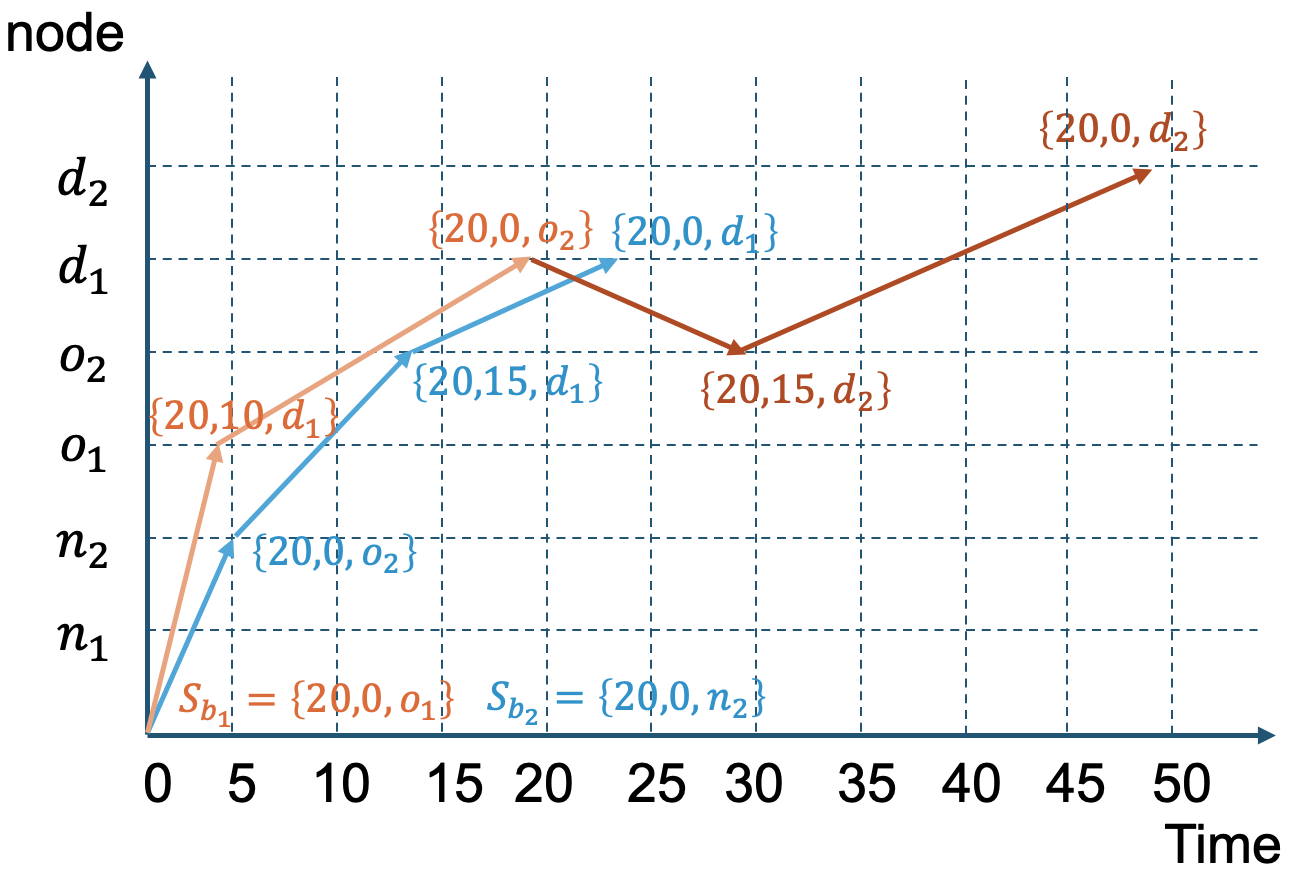}} \hspace{5pt}
    \caption{Space-time trajectory for feasible solutions. In this graph, the trajectories of Bus 1 are illustrated with orange lines and those of Bus 2 with blue lines. Different trips are represented by different shades of the same color. Each vector in the diagram is annotated with the status of all buses, represented as $s_b = \{c(b), p(b), d(b)\}$, where each parameter denotes the capacity of the bus, the number of evacuees on board and the next destination, respectively.}
    \label{Fig_feasible_solutions}
\end{figure*}

As shown in feasible solution 1 of Table \ref{tab:feasible_solutions} and Figure \ref{Fig_feasible_solutions} (a), Bus 1 begins its trip by departing from its initial location on link \( (n_1, o_1) \). At this point, the status of Bus 1 is represented as \( s_{b_1} = \{20, 0, o_1\} \), where \( c(b_1) = 20 \) indicates that the bus has a capacity of 20 passengers, \( p(b_1) = 0 \) means there are no passengers onboard, and \( d(b_1) = o_1 \) denotes that the next destination is the origin node \( o_1 \). Bus 1 takes 4 minutes to reach \( o_1 \), where it picks up 10 evacuees. After boarding the passengers, the status updates to \( s_{b_1} = \{20, 10, d_1\} \), meaning the bus now has 10 evacuees onboard, and its destination changes to shelter \( d_1 \). The bus then takes 15 minutes to transport these 10 evacuees from \( o_1 \) to \( d_1 \). The evacuees had waited 4 minutes for the bus, and their trip duration is 15 minutes, giving a total waiting + transportation time of 19 minutes per evacuee. After dropping off the passengers at \( d_1 \), the status of Bus 1 becomes \( s_{b_1} = \{20, 0, o_2\} \), as it now heads to the next origin \( o_2 \) with no passengers onboard. Subsequently, Bus 1 takes 10 minutes traveling to \( o_2 \), then 10 minutes transporting 15 passengers to the shelter \( d_1 \). These 15 evacuees waited a total of 29 minutes (19 minutes while Bus 1 was serving \( o_1 \), plus an additional 10 minutes for the trip to \( o_2 \)). Bus 1 then spends 10 minutes transporting these evacuees to \( d_1 \), after which its next destination continues to \(d_1\), which indicates this bus completes its service and is no longer dispatched to evacuate additional passengers.

Simultaneously, Bus 2 departs from its initial location and travels through \( n_2 \) towards the origin node \( o_2 \), since Bus 2 is located in the directed link \( (o_2, n_2) \).  Then Bus 2 evacuates 15 evacuees to shelter \( d_2 \). On this trip, the 15 evacuees wait 13 minutes before boarding, and Bus 2 takes 20 minutes to shelter \( d_2 \). Therefore, the total passenger time of feasible solution 1 is calculated by $19 \times 10 + 39 \times 15 + 33 \times 15 = 1,270$ minutes. If the penalty for inequity is $J_1$, the total reward of RL would be $1,270 + J_1$.

As shown in feasible solution 2 of Table \ref{tab:feasible_solutions} and Figure \ref{Fig_feasible_solutions} (b), Bus 1 first departs from its initial location to the origin node \(o_1\), then evacuates 10 passengers to shelter \(d_1\), contributing \((15 + 4) \times 10 = 190\) minutes of passenger time. Concurrently, Bus 2 departs from its initial location to origin node \(o_2\) and evacuates 15 passengers to shelter \(d_1\), adding \((20 + 29) \times 15 = 735\) minutes. After completing its first trip, Bus 1 proceeds to \( o_2 \) and transports the remaining 15 evacuees to shelter \( d_2 \), accruing another $(10 + 13) \times 15 = 345$ minutes. Thus, the total passenger time of this feasible solution is the same as in feasible solution 1, i.e., 1,270 minutes. The total reward for the RL agent is 1,270 + penalty $J_2$, where \(J_2\) accounts for the equity penalty.

Without accounting for equity, both solutions might seem equally optimal. However, when equity penalties \( J_1 \) and \( J_2 \) are considered, the prioritization of which community is evacuated first, or how resources are allocated, becomes crucial. In many real-world disaster scenarios, such as those in urban settings with diverse populations, vulnerable communities often require more immediate assistance. These are typically populations with limited mobility, lower socioeconomic status, or higher exposure to risk. In such cases, a strategy that serves these communities faster, even at the cost of slightly longer travel times for others, may be seen as more equitable. In this small-scale demonstration, the limited sample size and simplified network configuration prevent a meaningful calculation of the point-biserial coefficient and penalty $J$. In the San Francisco Bay Area network scenario discussed in the following sections, the diversity and size of the data allow for a more robust and statistically significant equity analysis.

\subsection{Optimal solution}\label{Optimal solution}
In this section, we introduce the use of the Equity-RL framework to simulate the six-node network and derive the optimal solution. 

The controller defines the policy for bus routing and evacuation decisions using the reinforcement learning approach, particularly the PPO algorithm.
\begin{enumerate}
    \item Once the bus is full, it will head to the nearest node with available capacity.
    \item Whenever the bus is not full, it will go to the nearest node with demand.
    \item The nearest node with demand is found using Dijkstra's algorithm.
    \item If there is no demand, but there are passengers on board, the bus will proceed to the nearest node with capacity.
\end{enumerate}

At each timestep, the PPO agent observes the current state and selects an action that specifies the next destination for each bus. The action space allows the agent to reroute buses to any node in the network. The PPO agent uses the current state to predict the next optimal actions, considering both efficiency and equity. Once an action is taken, the controller updates the state of the buses and the network. Each bus moves towards its selected destination, and the passengers are updated accordingly. When a bus arrives at its destination, evacuees either get on (if the bus is picking up passengers) or get off (if the bus is dropping them at a shelter). After this, the bus’s next destination is determined based on the current demand and capacity in the network, and the bus continues its route. The simulation continues until all evacuees have been evacuated or the maximum number of steps has been reached. The cumulative reward is then calculated, reflecting both the efficiency and equity of the evacuation.

\begin{figure*}[!ht]
    \centering
    \subfloat[Buses routing in Equity-RL framework]
    {\includegraphics[width=0.45\columnwidth]{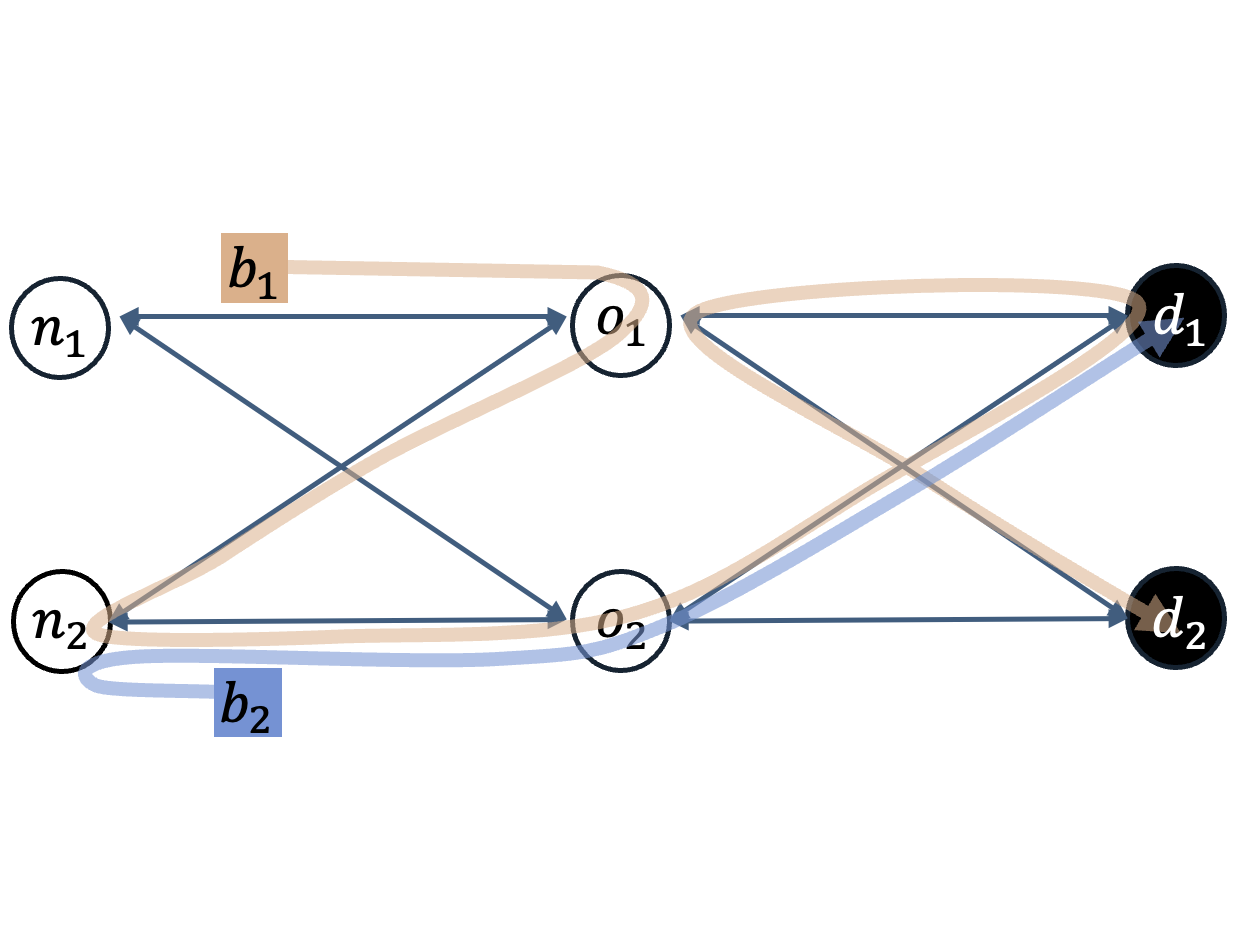}} \hspace{5pt}
    \subfloat[Space-time trajectory for buses]
    {\includegraphics[width=0.45\columnwidth]{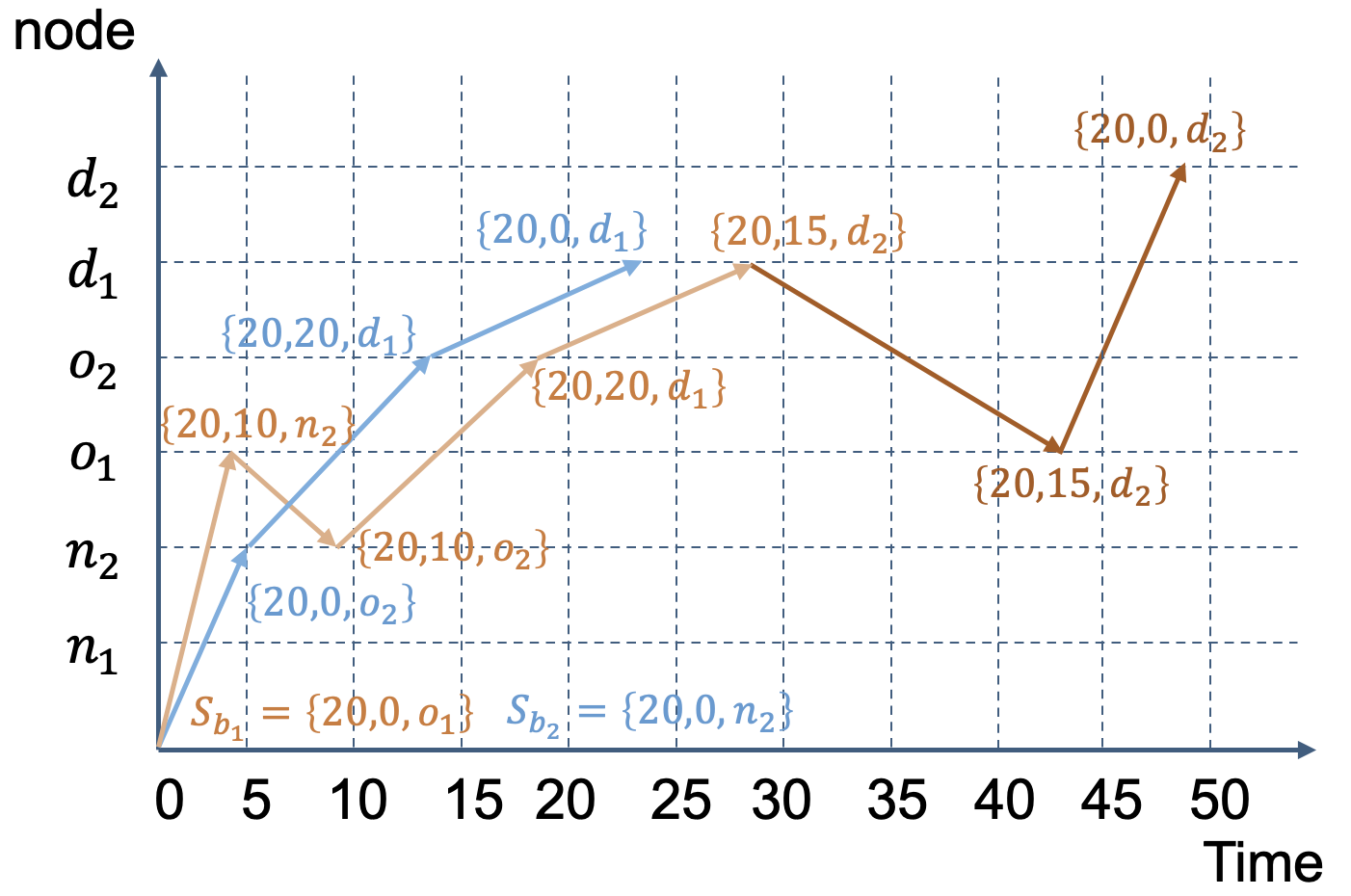}} \hspace{5pt}
    \caption{Optimal solution for six-node network}
    \label{Fig_optimal}
\end{figure*}

The reward is calculated based on a combination of total passenger time and inequity penalty. Here, for this illustrative network, we assume the penalty is the total passengers’ waiting time in inequitable communities. That is: $R = -(\text{total passenger time}) - (\text{total waiting time for passengers in } o_1)$. The objective is to maximize the reward value. As shown in Figure \ref{Fig_optimal} (b), the final output of the simulation includes bus real-time statues. Based on that, we can obtain the bus physical routing (refer to Figure \ref{Fig_optimal} (a)). For this illustrative case, the optimal reward is -1,345.

\section{Extended Simulation}
\label{sec:simulation}
Building on the insights from the simplified six-node network example, this section extends our model to a more complex real-world scenario. We apply our proposed evacuation strategies to a large-scale urban transportation network, San Francisco Bay Area, which allows us to test the effectiveness and scalability of the proposed equity-emphasized RL framework under more realistic and challenging conditions. 

\subsection{Data description}
\label{sec:data}

The success of the proposed approach relies heavily on accurate and comprehensive data. This section introduces the data required for the proposed system and its sources.

\subsubsection{Network data}
\begin{figure}[ht]
    \centering
    \includegraphics[width=0.6\columnwidth]{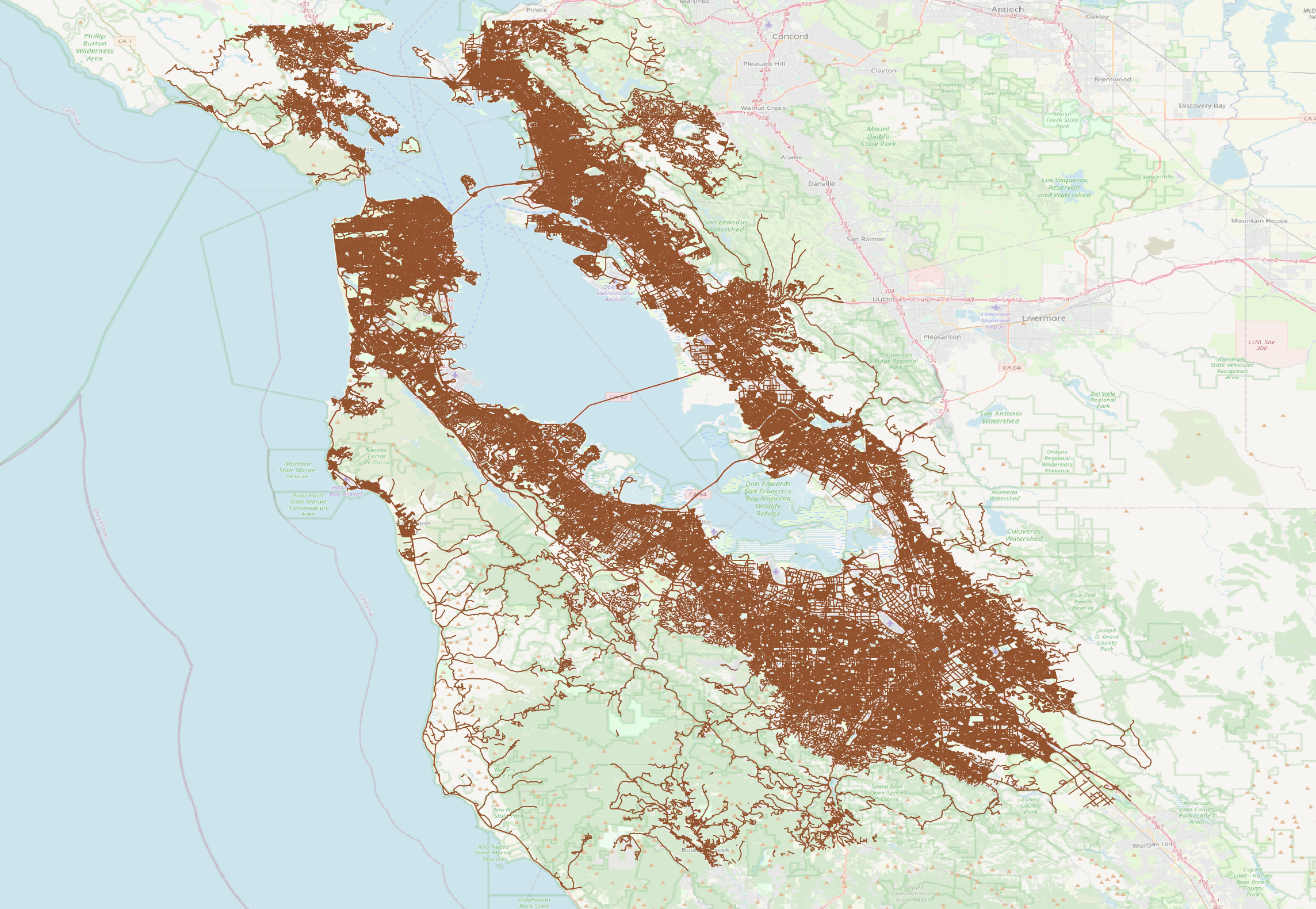}
    \caption{Visualization of the San Francisco Bay Area transportation network from \cite{OpenStreetMap2} .}
    \label{fig:osm_network}
\end{figure}

The transportation network is abstracted from various GIS-based data sources. In this study, the network is represented in GMNS format, and network information is provided by GTFS and OSM. The GTFS data provides real-time locations of buses and their schedules, while OSM data is utilized to extract the transportation network structure, including roads and intersections. Figure \ref{fig:osm_network} shows a visualization of the OSM-derived transportation network for the San Francisco Bay Area. The network consists of 416,677 links and 228,160 nodes, covering arterials, collectors, and local roads.

\subsubsection{Social data}

\begin{figure}[ht]
    \centering
    \includegraphics[width=0.6\columnwidth]{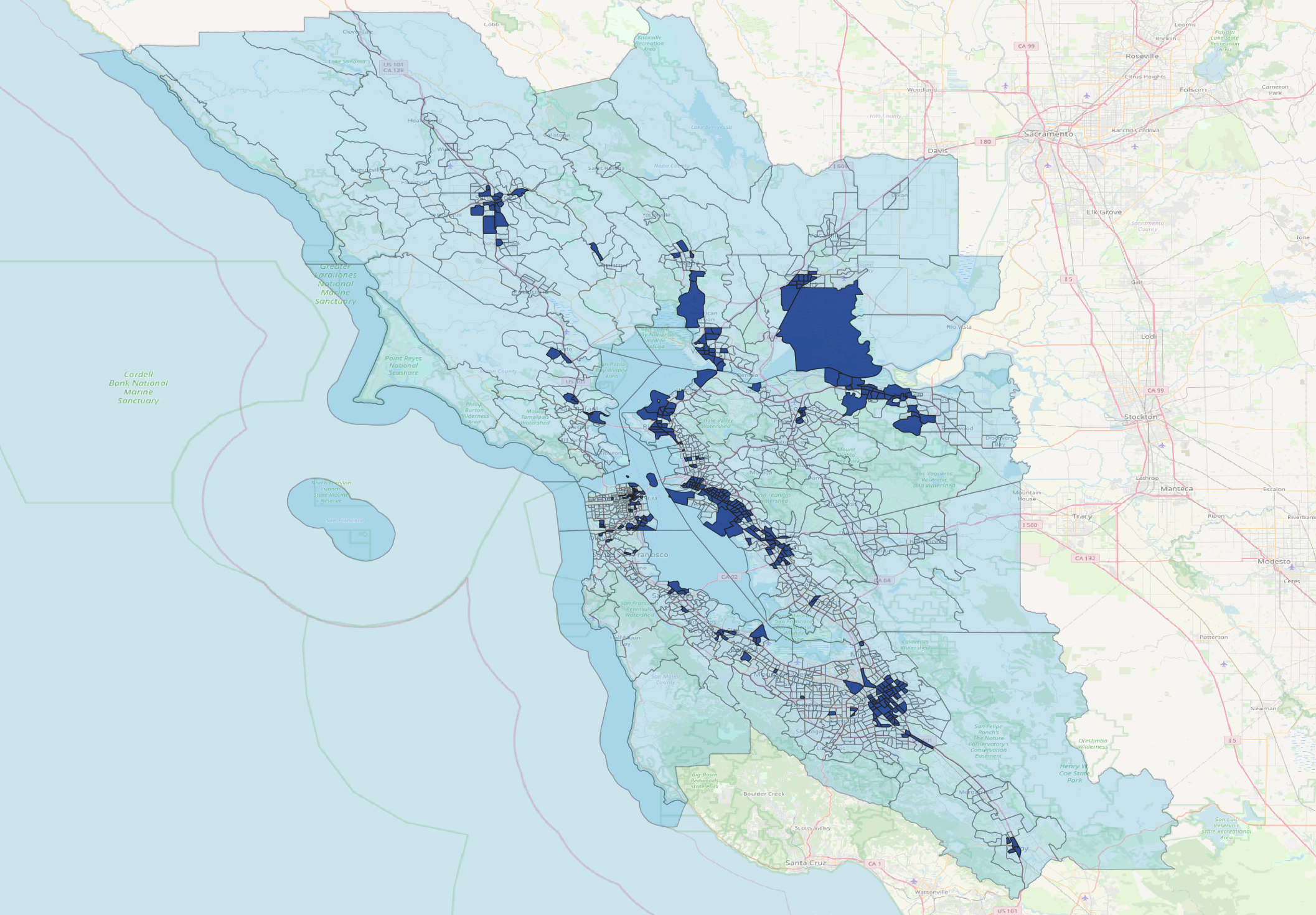}
    \caption{\cite{uscensus2018} division with equity priority communities highlighted.}
    \label{fig:census_tracts}
\end{figure}

Socio-demographic data is essential to generate traffic demand and identify EPCs. This study utilizes the U.S. Census 2018 data\cite{uscensus2018}, which provides detailed socio-demographic information. The geographical division for the analysis is based on U.S. Census tracts, as shown in Figure \ref{fig:census_tracts}. The EPC data, determined by the Metropolitan Transportation Commission (MTC), is used to identify equity priority communities.

\subsubsection{Natural hazards data}
To create realistic simulation scenarios, natural hazard data are required. The data curated by the Association of Bay Area Governments (ABAG) includes information on past wildfires, landslides, floods, and earthquakes. This data is used to simulate the impact of natural disasters on the transportation network. Table \ref{tab:hazards_data} summarizes the types of hazards and their corresponding data sources.

\begin{table}[ht]
\caption{Curated Data for Bay Area Hazards \citep{abag}}
\label{tab:hazards_data}
\begin{center}
\begin{tabular}{p{5cm}p{8cm}}
\toprule
\textbf{Hazard Type} & \textbf{Data Source} \\
\hline
Wildfire \citep{wildfire} & Fire and Resource Assessment Program (FRAP) \\
Landslide \citep{landslide} & US Geological Survey (USGS) \\
Flood \citep{flood} & Federal Emergency Management Agency (FEMA) \\
Earthquake \citep{earthquake} & California Geological Survey (CGS) \\
\bottomrule
\end{tabular}
\end{center}
\end{table}

\subsubsection{Real-time bus data}
\begin{figure}[ht]
    \centering
    \includegraphics[width=0.6\columnwidth]{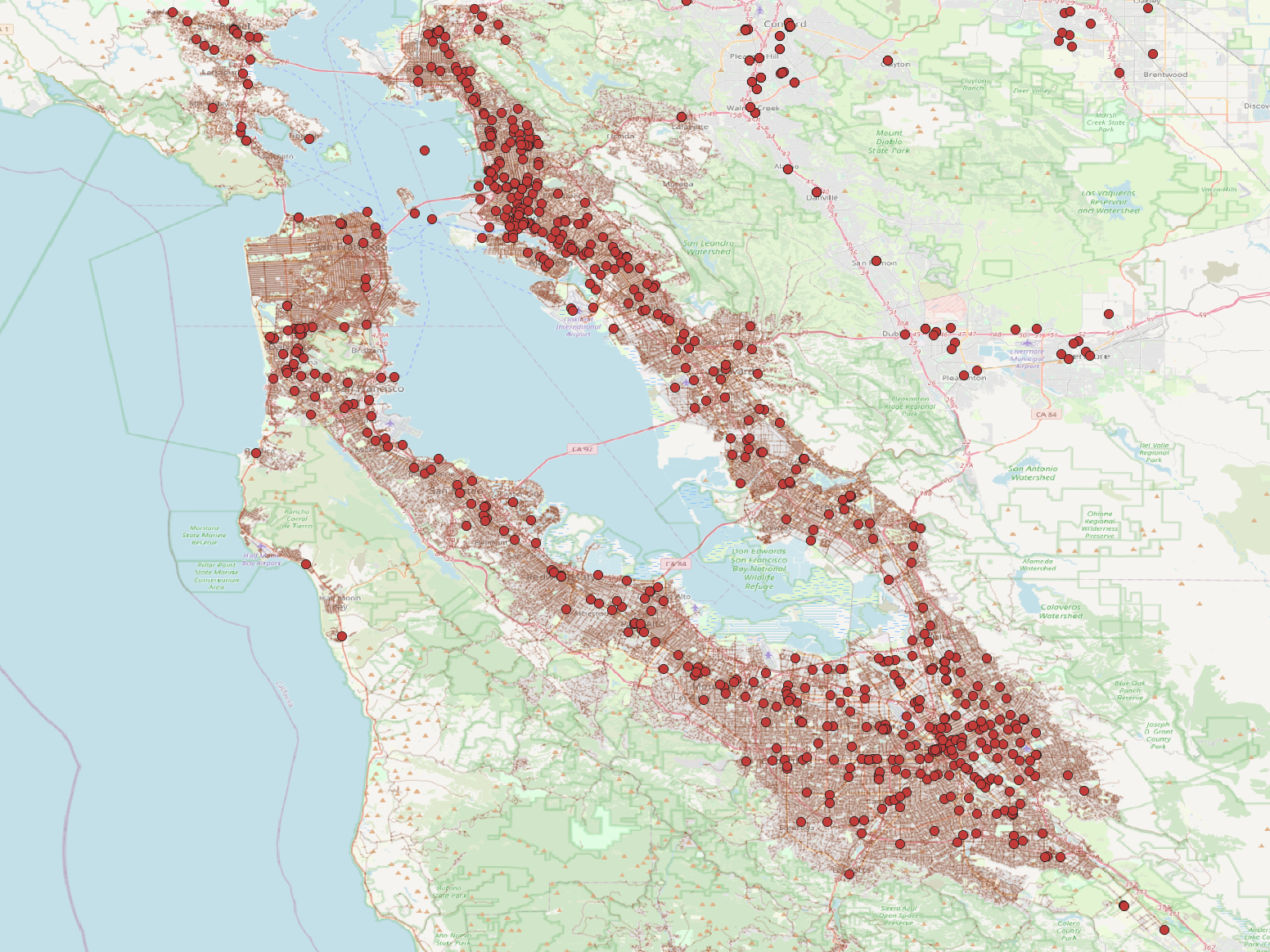}
    \caption{Spatial distribution of bus at 10 AM on Monday}
    \label{fig:bus_loc}
\end{figure}

The real-time bus data includes bus real-time location, which is extracted from GTFS data of twenty-two bus agencies (the list of bus agencies refer to the website https://511.org/transit/agencies).  Identifying bus geometries is based on bus schedule timings, stop geometry and the specific timestamp at which the bus location needs to be determined. If the bus coincides exactly with the physical location of a stop at the specified time (hazard time), then the geometry of the physical stop will be the bus geometry. If the bus is on the way from one stop to another stop, the bus location is determined by interpolating between known geographic locations of bus stops. Figure \ref{fig:bus_loc} represents the bus location at 10 AM on Monday. 

\subsection{Scenario generation}
\label{subsec:scenario_generation}

This section details the steps involved in creating realistic simulation scenarios by disabling links in affected zones, estimating evacuee populations, and initializing buses.

\begin{enumerate}
    \item \textbf{Determine Impairment Scale}: 
    The impairment scale refers to the extent of damage caused by a natural hazard. Historical hazard data is analyzed to extract the number of zones affected by past disasters. This data includes information on the frequency, intensity, and geographical spread of hazards such as wildfires, floods, landslides, and earthquakes. The impairment scale \(I\) is quantified as the number of zones impacted by a particular hazard event. This step ensures that the generated scenarios are grounded in empirical evidence and reflect realistic disaster conditions.

    \item \textbf{Select Impaired Zones}: 
    Once the impairment scale \(I\) is determined, the next step is to select the specific zones that will be impaired in the simulation. Two types of scenarios are considered:
    \begin{itemize}
        \item \textit{Hazard Reproduce Scenarios}: In these scenarios, impaired zones are directly mapped from historical hazard datasets. This involves identifying the geographical areas affected by previous disasters and replicating those conditions in the simulation.
        \item \textit{Randomized Scenarios}: In these scenarios, zones are randomly selected based on the impairment scale \(I\). This approach simulates the spatial distribution of damage in a more stochastic manner, allowing for the assessment of evacuation strategies under varying conditions. The random selection process is guided by a probability distribution that reflects the likelihood of different zones being affected.
    \end{itemize}

    \item \textbf{Evacuee Estimation}: 
    In the selected impaired zones, all affiliated nodes (such as residential areas, schools and hospitals) are marked as origin nodes for evacuees. The total population \(P_i\) in each impaired zone \(i\) is obtained from demographic data sources, such as census data. The population is then distributed equally among the affiliated nodes \(N_i\), resulting in the number of evacuees \(E_{ij}\) at each node \(j\) in zone \(i\):
    \begin{equation}
    E_{ij} = \frac{P_i}{N_i}
    \end{equation}
    This distribution assumes a uniform spread of evacuees across nodes, which can be adjusted based on more detailed population density data if available.

    \item \textbf{Bus Initialization}: 
    Buses are initialized based on real-time data from the GTFS data. The GTFS provides detailed information on the location, capacity, and schedule of buses. The initialization process involves the following steps:
    \begin{itemize}
        \item \textit{Location Mapping}: The initial locations of buses are mapped according to their positions at the time of day when the hazard event starts. This involves assigning each bus to a specific node or link in the transportation network.
        \item \textit{Capacity Setting}: Each bus is assigned a capacity \(C_b\), representing the maximum number of evacuees it can transport. The initial number of passengers on board \(P_b\) is also set, which may vary depending on the bus's current route and occupancy.
        \item \textit{Route Initialization}: The initial routes of buses are established based on their schedules and the current state of the transportation network. This includes accounting for any disruptions caused by the hazard event, such as disabled links or road closures.
    \end{itemize}
    This detailed initialization ensures that the simulation starts from a realistic state, reflecting the actual conditions of the transportation network and the bus fleet.

\end{enumerate}

By following these steps, the scenario generation process creates varied conditions under which the proposed evacuation strategies can be tested. This approach allows for the assessment of both the efficiency and equity of the strategies in different disaster scenarios, providing insights into their robustness and applicability in real-world situations.

\subsection{Benchmarks}

To evaluate the performance of the proposed RL-based strategy, we compare it with the following control strategies:
\begin{itemize}
    \item \textbf{Stochastic Strategy 1 (\( \pi_1 \))}: Randomly assigns buses to destinations from all impaired or shelter nodes.
    \item \textbf{Stochastic Strategy 2 (\( \pi_2 \))}: Randomly assigns buses to destinations only from the impaired zones with waiting evacuees and shelters with available capacity.
    \item \textbf{Rule-based Strategy 1 (\( \bar{\pi}_1 \))}: Prioritizes bus assignments based on the shortest travel time to impaired zones and shelters.
    \item \textbf{Rule Strategy 2 (\( \bar{\pi}_2 \))}: Prioritizes bus assignments based on the number of evacuees at the impaired nodes.
    \item \textbf{Efficiency-emphasized RL Strategy}: The RL strategy with all the same settings without equity penalty in the reward function.
\end{itemize}

The performance of these strategies is evaluated based on the reward function, focusing on both efficiency and equity. 

\section{Results}
\label{sec:results}

\begin{table*}[!h]
\caption{Comparison of Evacuation Strategies}
\label{tab:comparison_results}
\begin{center}
\begin{tabular}{p{3.8cm}p{1.8cm}p{1.2cm}p{1.2cm}p{1.2cm}p{1.2cm}p{1.2cm}p{1.2cm}}
\toprule
\textbf{Strategy} & \textbf{Total Evacuation Time ($10^3$min)} & \textbf{Overall \( |r_{pb}| \)} & \textbf{Wildfire \( |r_{pb}| \)} &  \textbf{Landsilde \( |r_{pb}| \)} & \textbf{Flood \( |r_{pb}| \)} & \textbf{Earthquake \( |r_{pb}| \)} & \textbf{Random \( |r_{pb}| \)}\\
\hline
Equity-RL Strategy & 1395 & 0.102& 0.089& 0.063& 0.118& 0.109& 0.097 \\
Efficiency-RL Strategy & 1281 & 0.189& 0.170& 0.178& 0.144& 0.262& 0.209 \\
Random Strategy 1 (\( \pi_1 \)) & 6437 & 0.156& 0.165& 0.108& 0.196& 0.131& 0.110\\
Random Strategy 2 (\( \pi_2 \)) & 2908 & 0.172& 0.104& 0.181& 0.143& 0.195& 0.169\\
Rule-based Strategy 1 (\( \bar{\pi}_1 \)) & 1496 & 0.191& 0.223& 0.207& 0.164& 0.196& 0.153\\
Rule-based Strategy 2 (\( \bar{\pi}_2 \)) & 1712 & 0.169& 0.154& 0.189& 0.176& 0.127& 0.184\\
\bottomrule
\end{tabular}
\end{center}
\end{table*}

This section presents a comprehensive analysis of the simulation results obtained from the implementation of the proposed Equity-Emphasized Reinforcement Learning (Equity-RL) strategy for bus evacuation planning. The performance of this strategy is evaluated against several benchmarks, including an Efficiency-Emphasized RL strategy, two stochastic strategies, and two rule-based strategies. The key metrics used for this comparison are the total evacuation time, the overall equity index, as measured by the point-biserial correlation coefficient ($|r_{pb}|$), and equity indices specific to hazard types such as wildfires, landslides, floods, and earthquakes. Table \ref{tab:comparison_results} summarizes the results of these strategies.

\subsection{Overall evacuation time analysis}
The total evacuation time provides a direct measure of how efficiently each strategy manages the evacuation process. As shown in Table \ref{tab:comparison_results}, the Equity-RL strategy achieves a total evacuation time of 1,395,000 minutes. This represents a balance between minimizing the evacuation duration and maintaining an equitable service distribution across affected areas. In comparison, the Efficiency-RL strategy, which prioritizes evacuation speed without explicit equity considerations, achieves a shorter total evacuation time of 1,281,000 minutes. This result demonstrates the trade-off between efficiency and equity, where faster evacuation times can come at the expense of fair service distribution.

Stochastic Strategy 1 (\(\pi_1\)) recorded the longest evacuation time at 6,437,000 minutes, followed by Stochastic Strategy 2 (\(\pi_2\)) at 2,908,000 minutes. These results indicate that random bus assignments lead to significant inefficiencies, likely due to suboptimal routing and scheduling decisions. The rule-based strategies performed better than the stochastic ones, with Rule-based Strategy 1 (\(\bar{\pi}_1\)) achieving an evacuation time of 1,496,000 minutes and Rule-based Strategy 2 (\(\bar{\pi}_2\)) at 1,712,000 minutes. While these strategies offer improvements over random assignments, they are still outperformed by both RL-based approaches in terms of total evacuation time.

\subsection{Equity index analysis}
The overall equity index, as measured by the point-biserial correlation coefficient ($|r_{pb}|$), provides an important metric for assessing how equitably the evacuation resources are distributed across different communities, especially those identified as vulnerable. The Equity-RL strategy achieved the lowest overall equity index of 0.102, indicating the most equitable resource distribution among all strategies. This performance suggests that the RL agent was able to effectively prioritize EPC during the evacuation process.

In contrast, the Efficiency-RL strategy recorded an overall equity index of 0.189, which reflects the trade-off made for faster evacuation times. This higher inequity index suggests that the Efficiency-RL strategy tended to favor more easily accessible areas over vulnerable communities, resulting in less equitable resource distribution.

The stochastic strategies also demonstrated higher inequity indices, with Strategy 1 (\(\pi_1\)) achieving an equity index of 0.156 and Strategy 2 (\(\pi_2\)) recording 0.172. These values indicate that random bus assignments fail to account for the specific needs of vulnerable populations, leading to less equitable outcomes. The rule-based strategies showed even higher inequity, with Rule-based Strategy 1 (\(\bar{\pi}_1\)) and Rule-based Strategy 2 (\(\bar{\pi}_2\)) recording indices of 0.191 and 0.169, respectively.

\subsection{Hazard-specific equity analysis}
To further understand the performance of the proposed strategies across different disaster scenarios, hazard-specific equity indices were calculated for wildfires, landslides, floods, and earthquakes. The Equity-RL strategy consistently achieved the lowest inequity indices across all hazard types, highlighting its robustness in prioritizing vulnerable communities regardless of the specific nature of the disaster.

For instance, in wildfire scenarios, the Equity-RL strategy recorded an equity index of 0.089, which is significantly lower than the 0.170 observed with the Efficiency-RL strategy. This indicates that, under wildfire conditions, the Equity-RL strategy was more effective in ensuring that equity priority communities received adequate attention. Similarly, in landslide scenarios, the Equity-RL strategy achieved an equity index of 0.063, compared to 0.178 for the Efficiency-RL strategy, demonstrating its superior performance in addressing equity under landslide-related evacuations.

In flood scenarios, the Equity-RL strategy recorded an equity index of 0.118, again outperforming the Efficiency-RL strategy, which recorded 0.144. Lastly, for earthquake scenarios, the Equity-RL strategy's equity index was 0.109, while the Efficiency-RL strategy recorded a much higher index of 0.262. These results underscore the ability of the Equity-RL strategy to adapt to various disaster types and ensure that vulnerable populations are not neglected in the evacuation process.

\subsection{Evacuation time distribution analysis}
\begin{figure}[ht]
    \centering
    \includegraphics[width=0.8\linewidth]{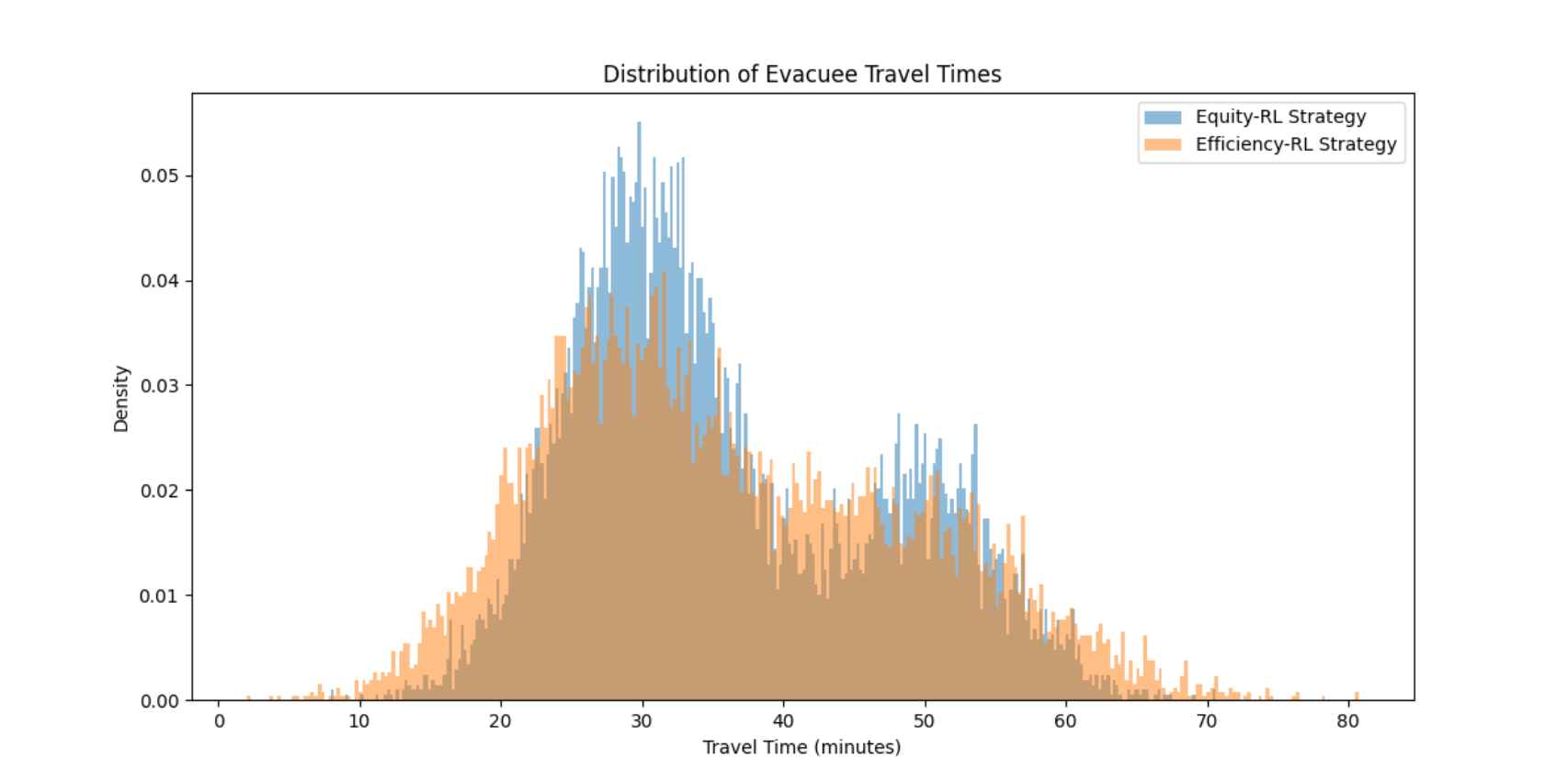}
    \caption{Distribution of evacuee travel times for Equity-RL and Efficiency-RL strategies.}
    \label{fig:evacuee_travel_times}
\end{figure}

\begin{figure}[ht]
    \centering
    \includegraphics[width=0.8\linewidth]{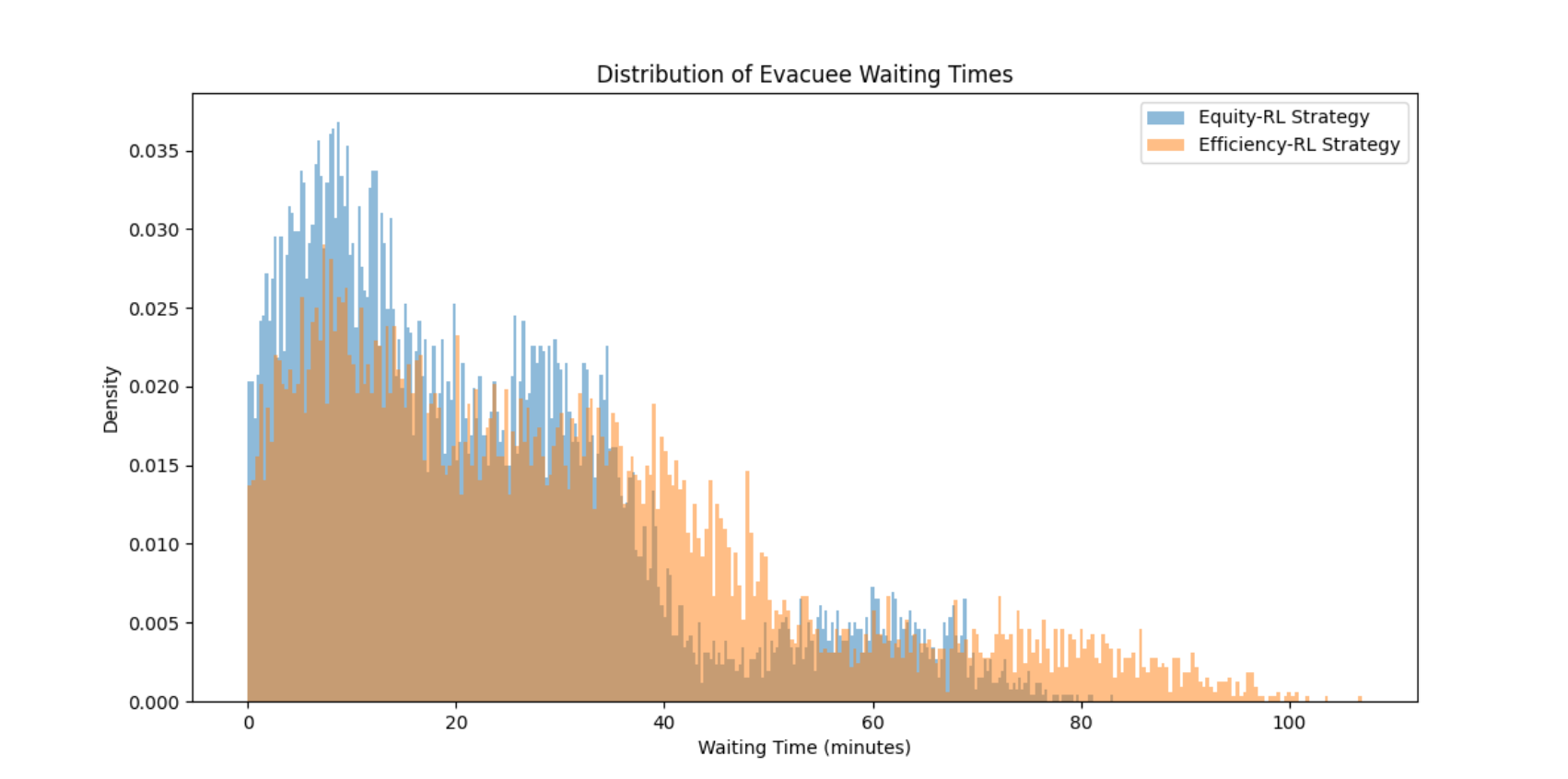}
    \caption{Distribution of evacuee waiting times for Equity-RL and Efficiency-RL strategies.}
    \label{fig:evacuee_waiting_times}
\end{figure}

To provide further insight into the distribution of evacuation times, we generated histograms of individual evacuee travel times and waiting times for both the Equity-RL and Efficiency-RL strategies. These distributions are illustrated in Figures \ref{fig:evacuee_travel_times} and \ref{fig:evacuee_waiting_times}, which allow for a direct comparison of the variability in travel and waiting times across the two strategies.

Figure \ref{fig:evacuee_travel_times} shows the distribution of travel times for evacuees under both strategies. The travel times for both the Equity-RL and Efficiency-RL strategies are centered around a similar range, with the peak occurring between 25 and 30 minutes. 

\begin{itemize}
    \item \textbf{Equity-RL Strategy}: The travel time distribution for the Equity-RL strategy, represented by the blue curve, is more concentrated around this peak. This indicates that a larger proportion of evacuees experienced travel times within this specific range, suggesting a more uniform distribution of travel times across evacuees. 
    \item \textbf{Efficiency-RL Strategy}: The Efficiency-RL strategy, shown by the orange curve, exhibits a wider spread in travel times, particularly with a longer tail extending beyond 40 minutes and up to approximately 80 minutes. This suggests that while many evacuees had travel times comparable to the Equity-RL strategy, a small percentage of evacuees experienced significantly longer travel times.
\end{itemize}

The key difference between the two strategies lies in the consistency of travel times. The Equity-RL strategy maintains a tighter distribution, minimizing the number of evacuees experiencing travel times beyond 50 minutes. In contrast, the Efficiency-RL strategy allows for greater variability, with some evacuees experiencing travel times as high as 80 minutes.

Figure \ref{fig:evacuee_waiting_times} presents the distribution of waiting times for both strategies. 

\begin{itemize}
    \item \textbf{Equity-RL Strategy}: The Equity-RL strategy again demonstrates a more concentrated distribution of waiting times, with the peak occurring between 20 and 30 minutes. The distribution of waiting times shows that a significant portion of evacuees waited between 10 and 40 minutes, with a small number of evacuees experiencing longer waiting times beyond 50 minutes.
    \item \textbf{Efficiency-RL Strategy}: The Efficiency-RL strategy, on the other hand, displays a broader distribution with a lower peak. A notable difference is the longer tail, with many evacuees experiencing waiting times exceeding 50 minutes and some reaching up to 100 minutes. This suggests that while the strategy minimizes total evacuation time, certain evacuees, particularly those in equity priority communities, are subjected to longer waiting times.
\end{itemize}

The Equity-RL strategy leads to fewer extreme cases of long waiting times, contributing to a more equitable distribution of service. The Efficiency-RL strategy has a larger variance in waiting times, with more evacuees experiencing extended delays in bus arrival and service.

The histograms demonstrate the trade-offs between the two strategies in terms of equity and efficiency:
\begin{itemize}
    \item The Equity-RL strategy exhibits more uniform and concentrated distributions for both travel and waiting times. This strategy ensures that evacuees have a more consistent evacuation experience, reducing the number of evacuees who experience particularly long travel or waiting times. As a result, the Equity-RL strategy better addresses the equity concerns, ensuring that vulnerable communities are not disproportionately affected by long waits or extended travel times.
    \item The Efficiency-RL strategy, while minimizing the total evacuation time, introduces greater variability in both waiting and travel times. This variability can result in some evacuees, particularly from equity priority areas, experiencing significantly longer evacuation times. The long tail observed in both distributions suggests that certain evacuees endure considerably longer delays.
\end{itemize}

Overall, the Equity-RL strategy prioritizes equitable service delivery across all evacuees, leading to a more balanced evacuation process. In contrast, the Efficiency-RL strategy, while faster overall, results in less consistent outcomes, potentially exacerbating inequalities in service access during emergencies.

\section{Conclusion}
\label{sec:conclusion}

This study introduced and evaluated an Equity-RL framework for bus evacuation planning. The results demonstrate that this framework offers a viable solution for balancing efficiency and equity in bus evacuation planning, achieving reasonable evacuation times while maintaining a low inequity index across various disaster scenarios. While efficiency-focused and traditional frameworks achieve shorter evacuation times, they do so at the cost of equitable service distribution, particularly for vulnerable communities. The comparison with the traditional Efficiency-RL framework further demonstrates the superior performance of the proposed Equity-RL framework which contributes meaningful insights into the optimization of evacuation processes. This study underscores the importance of integrating advanced computational techniques with social equity principles to develop more effective and fair emergency management practices. Future research should focus on extending these models to larger and more complex scenarios, incorporating dynamic real-time data, and exploring the integration of multiple modes of transportation to optimize evacuation strategies for diverse urban environments further.

% \clearpage %%Remove this from your manuscript

% \appendix
% \section{Appendix}

% % To print the credit authorship contribution details

\subsubsection*{Declaration of generative AI and AI-assisted technologies in the writing process}
\noindent During the preparation of this work ChatGPT was used to refine the language and clarity of text written by non-native speakers. After using this tool/service, the authors reviewed and edited the content as needed and take  full responsibility for the content of the publication.

\bibliographystyle{cas-model2-names}

\bibliography{cas-sc-template.bbl}

\begin{thebibliography}{30}
\expandafter\ifx\csname natexlab\endcsname\relax\def\natexlab#1{#1}\fi
\providecommand{\url}[1]{\texttt{#1}}
\providecommand{\href}[2]{#2}
\providecommand{\path}[1]{#1}
\providecommand{\DOIprefix}{doi:}
\providecommand{\ArXivprefix}{arXiv:}
\providecommand{\URLprefix}{URL: }
\providecommand{\Pubmedprefix}{pmid:}
\providecommand{\doi}[1]{\href{http://dx.doi.org/#1}{\path{#1}}}
\providecommand{\Pubmed}[1]{\href{pmid:#1}{\path{#1}}}
\providecommand{\bibinfo}[2]{#2}
\ifx\xfnm\relax \def\xfnm[#1]{\unskip,\space#1}\fi
%Type = Article
\bibitem[{Abdelgawad et~al.(2010)Abdelgawad, Abdulhai and Wahba}]{abdelgawad2010multiobjective}
\bibinfo{author}{Abdelgawad, H.}, \bibinfo{author}{Abdulhai, B.}, \bibinfo{author}{Wahba, M.}, \bibinfo{year}{2010}.
\newblock \bibinfo{title}{Multiobjective optimization for multimodal evacuation}.
\newblock \bibinfo{journal}{Transportation Research Record} \bibinfo{volume}{2196}, \bibinfo{pages}{21--33}.
%Type = Misc
\bibitem[{Agency()}]{flood}
\bibinfo{author}{Agency, F.E.M.}, .
\newblock \bibinfo{title}{{FEMA Flood Maps}}.
\newblock \URLprefix \url{https://www.fema.gov/flood-maps}. \bibinfo{note}{(2021, Oct 10)}.
%Type = Article
\bibitem[{Altay and Green~III(2006)}]{altay2006or}
\bibinfo{author}{Altay, N.}, \bibinfo{author}{Green~III, W.G.}, \bibinfo{year}{2006}.
\newblock \bibinfo{title}{Or/ms research in disaster operations management}.
\newblock \bibinfo{journal}{European journal of operational research} \bibinfo{volume}{175}, \bibinfo{pages}{475--493}.
%Type = Misc
\bibitem[{from Association~of Bay Area~Governments()}]{abag}
\bibinfo{author}{from Association~of Bay Area~Governments}, .
\newblock \bibinfo{title}{{Curated Data for Bay Area Hazards}}.
\newblock \URLprefix \url{https://abag.ca.gov/our-work/resilience/data-research}. \bibinfo{note}{(2021, Oct 10)}.
%Type = Article
\bibitem[{Bish(2011)}]{bish2011planning}
\bibinfo{author}{Bish, D.R.}, \bibinfo{year}{2011}.
\newblock \bibinfo{title}{Planning for a bus-based evacuation}.
\newblock \bibinfo{journal}{OR spectrum} \bibinfo{volume}{33}, \bibinfo{pages}{629--654}.
%Type = Article
\bibitem[{Chen and Chou(2009)}]{chen2009modeling}
\bibinfo{author}{Chen, C.C.}, \bibinfo{author}{Chou, C.S.}, \bibinfo{year}{2009}.
\newblock \bibinfo{title}{Modeling and performance assessment of a transit-based evacuation plan within a contraflow simulation environment}.
\newblock \bibinfo{journal}{Transportation Research Record} \bibinfo{volume}{2091}, \bibinfo{pages}{40--50}.
%Type = Article
\bibitem[{Chen and Zhan(2008)}]{chen2008agent}
\bibinfo{author}{Chen, X.}, \bibinfo{author}{Zhan, F.B.}, \bibinfo{year}{2008}.
\newblock \bibinfo{title}{Agent-based modelling and simulation of urban evacuation: relative effectiveness of simultaneous and staged evacuation strategies}.
\newblock \bibinfo{journal}{Journal of the Operational Research Society} \bibinfo{volume}{59}, \bibinfo{pages}{25--33}.
%Type = Article
\bibitem[{Cova and Johnson(2003)}]{cova2003network}
\bibinfo{author}{Cova, T.J.}, \bibinfo{author}{Johnson, J.P.}, \bibinfo{year}{2003}.
\newblock \bibinfo{title}{A network flow model for lane-based evacuation routing}.
\newblock \bibinfo{journal}{Transportation research part A: Policy and Practice} \bibinfo{volume}{37}, \bibinfo{pages}{579--604}.
%Type = Misc
\bibitem[{FIRE()}]{wildfire}
\bibinfo{author}{FIRE, C.}, .
\newblock \bibinfo{title}{{The Fire and Resource Assessment Program (FRAP) - Fire Perimeters}}.
\newblock \URLprefix \url{https://frap.fire.ca.gov/frap-projects/fire-perimeters/}. \bibinfo{note}{(2021, Oct 10)}.
%Type = Misc
\bibitem[{{General Transit Feed Specification (GTFS)}(2024)}]{gtfs_data_transitfeeds}
\bibinfo{author}{{General Transit Feed Specification (GTFS)}}, \bibinfo{year}{2024}.
\newblock \bibinfo{title}{General transit feed specification (gtfs) data}.
\newblock \URLprefix \url{https://transitfeeds.com/}. \bibinfo{note}{accessed: 2024-11-01}.
%Type = Article
\bibitem[{Goerigk and Gr{\"u}n(2014)}]{goerigk2014robust}
\bibinfo{author}{Goerigk, M.}, \bibinfo{author}{Gr{\"u}n, B.}, \bibinfo{year}{2014}.
\newblock \bibinfo{title}{A robust bus evacuation model with delayed scenario information}.
\newblock \bibinfo{journal}{Or Spectrum} \bibinfo{volume}{36}, \bibinfo{pages}{923--948}.
%Type = Article
\bibitem[{Goerigk et~al.(2013)Goerigk, Gr{\"u}n and He{\ss}ler}]{goerigk2013branch}
\bibinfo{author}{Goerigk, M.}, \bibinfo{author}{Gr{\"u}n, B.}, \bibinfo{author}{He{\ss}ler, P.}, \bibinfo{year}{2013}.
\newblock \bibinfo{title}{Branch and bound algorithms for the bus evacuation problem}.
\newblock \bibinfo{journal}{Computers \& Operations Research} \bibinfo{volume}{40}, \bibinfo{pages}{3010--3020}.
%Type = Techreport
\bibitem[{Hamacher and Tjandra(2001)}]{HamacherTjandra2001}
\bibinfo{author}{Hamacher, H.}, \bibinfo{author}{Tjandra, S.}, \bibinfo{year}{2001}.
\newblock \bibinfo{title}{Mathematical Modelling of Evacuation Problems: A State of Art}.
\newblock \bibinfo{type}{Technical Report} \bibinfo{number}{24}. Fraunhofer (ITWM).
\newblock \URLprefix \url{https://nbn-resolving.de/urn:nbn:de:hbz:386-kluedo-12873}.
%Type = Article
\bibitem[{Huang et~al.(2016)Huang, Lindell and Prater}]{huang2016leaves}
\bibinfo{author}{Huang, S.K.}, \bibinfo{author}{Lindell, M.K.}, \bibinfo{author}{Prater, C.S.}, \bibinfo{year}{2016}.
\newblock \bibinfo{title}{Who leaves and who stays? a review and statistical meta-analysis of hurricane evacuation studies}.
\newblock \bibinfo{journal}{Environment and Behavior} \bibinfo{volume}{48}, \bibinfo{pages}{991--1029}.
%Type = Article
\bibitem[{Kaiser et~al.(2012)Kaiser, Hess and Palomo}]{kaiser2012emergency}
\bibinfo{author}{Kaiser, E.I.}, \bibinfo{author}{Hess, L.}, \bibinfo{author}{Palomo, A.B.P.}, \bibinfo{year}{2012}.
\newblock \bibinfo{title}{An emergency evacuation planning model for special needs populations using public transit systems}.
\newblock \bibinfo{journal}{Journal of Public Transportation} \bibinfo{volume}{15}, \bibinfo{pages}{45--69}.
%Type = Article
\bibitem[{Khalili et~al.(2024)Khalili, Mojtahedi, Steinmetz-Weiss and Sanderson}]{khalili2024systematic}
\bibinfo{author}{Khalili, S.M.}, \bibinfo{author}{Mojtahedi, M.}, \bibinfo{author}{Steinmetz-Weiss, C.}, \bibinfo{author}{Sanderson, D.}, \bibinfo{year}{2024}.
\newblock \bibinfo{title}{A systematic literature review on transit-based evacuation planning in emergency logistics management: Optimisation and modelling approaches}.
\newblock \bibinfo{journal}{Buildings} \bibinfo{volume}{14}, \bibinfo{pages}{176}.
%Type = Article
\bibitem[{Litman(2006)}]{litman2006lessons}
\bibinfo{author}{Litman, T.}, \bibinfo{year}{2006}.
\newblock \bibinfo{title}{Lessons from katrina and rita: What major disasters can teach transportation planners}.
\newblock \bibinfo{journal}{Journal of transportation engineering} \bibinfo{volume}{132}, \bibinfo{pages}{11--18}.
%Type = Article
\bibitem[{Liu et~al.(2006)Liu, Lai and Chang}]{liu2006two}
\bibinfo{author}{Liu, Y.}, \bibinfo{author}{Lai, X.}, \bibinfo{author}{Chang, G.L.}, \bibinfo{year}{2006}.
\newblock \bibinfo{title}{Two-level integrated optimization system for planning of emergency evacuation}.
\newblock \bibinfo{journal}{Journal of transportation Engineering} \bibinfo{volume}{132}, \bibinfo{pages}{800--807}.
%Type = Inproceedings
\bibitem[{Lu et~al.(2005)Lu, George and Shekhar}]{lu2005capacity}
\bibinfo{author}{Lu, Q.}, \bibinfo{author}{George, B.}, \bibinfo{author}{Shekhar, S.}, \bibinfo{year}{2005}.
\newblock \bibinfo{title}{Capacity constrained routing algorithms for evacuation planning: A summary of results}, in: \bibinfo{booktitle}{International symposium on spatial and temporal databases}, \bibinfo{organization}{Springer}. pp. \bibinfo{pages}{291--307}.
%Type = Misc
\bibitem[{{Metropolitan Transportation Commission}(2021)}]{epc}
\bibinfo{author}{{Metropolitan Transportation Commission}}, \bibinfo{year}{2021}.
\newblock \bibinfo{title}{Mtc plan bay area 2050 equity priority communities}.
\newblock \URLprefix \url{https://github.com/BayAreaMetro/Spatial-Analysis-Mapping-Projects/tree/master/Project-Documentation/Equity-Priority-Communities}.
%Type = Misc
\bibitem[{{OpenStreetMap}(2024)}]{OpenStreetMap2}
\bibinfo{author}{{OpenStreetMap}}, \bibinfo{year}{2024}.
\newblock \bibinfo{title}{Openstreetmap: Geospatial data for mapping}.
\newblock \URLprefix \url{https://www.openstreetmap.org/#map=10/33.6049/-112.1248}. \bibinfo{note}{accessed: 2024-11-01}.
%Type = Misc
\bibitem[{{OpenStreetMap (OSM)}(2024)}]{OpenStreetMap}
\bibinfo{author}{{OpenStreetMap (OSM)}}, \bibinfo{year}{2024}.
\newblock \bibinfo{title}{Openstreetmap: Geospatial data for mapping}.
\newblock \URLprefix \url{https://www.openstreetmap.org/#map=10/33.6049/-112.1248}. \bibinfo{note}{accessed: 2024-11-01}.
%Type = Article
\bibitem[{Renne et~al.(2011)Renne, Sanchez and Litman}]{renne2011carless}
\bibinfo{author}{Renne, J.L.}, \bibinfo{author}{Sanchez, T.W.}, \bibinfo{author}{Litman, T.}, \bibinfo{year}{2011}.
\newblock \bibinfo{title}{Carless and special needs evacuation planning: A literature review}.
\newblock \bibinfo{journal}{Journal of Planning Literature} \bibinfo{volume}{26}, \bibinfo{pages}{420--431}.
%Type = Article
\bibitem[{Sayyady and Eksioglu(2010)}]{sayyady2010optimizing}
\bibinfo{author}{Sayyady, F.}, \bibinfo{author}{Eksioglu, S.D.}, \bibinfo{year}{2010}.
\newblock \bibinfo{title}{Optimizing the use of public transit system during no-notice evacuation of urban areas}.
\newblock \bibinfo{journal}{Computers \& Industrial Engineering} \bibinfo{volume}{59}, \bibinfo{pages}{488--495}.
%Type = Misc
\bibitem[{Survey()}]{earthquake}
\bibinfo{author}{Survey, C.G.}, .
\newblock \bibinfo{title}{{Hazards Program, 2018}}.
\newblock \URLprefix \url{https://www.conservation.ca.gov/cgs/shp}. \bibinfo{note}{(2021, Oct 10)}.
%Type = Misc
\bibitem[{{U.S. Census Bureau}(2018)}]{uscensus2018}
\bibinfo{author}{{U.S. Census Bureau}}, \bibinfo{year}{2018}.
\newblock \bibinfo{title}{2018 american community survey 1-year estimates}.
\newblock \URLprefix \url{https://data.census.gov/cedsci/table?q=2018%20ACS%201-Year%20Estimates}. \bibinfo{note}{accessed: 2024-11-01}.
%Type = Inproceedings
\bibitem[{Wang and Delle~Monache(2022)}]{wang2022urban}
\bibinfo{author}{Wang, H.}, \bibinfo{author}{Delle~Monache, M.L.}, \bibinfo{year}{2022}.
\newblock \bibinfo{title}{Urban network resilience analysis and equity emphasized recovery based on reinforcement learning}, in: \bibinfo{booktitle}{2022 European Control Conference (ECC)}, \bibinfo{organization}{IEEE}. pp. \bibinfo{pages}{01--06}.
%Type = Techreport
\bibitem[{Wentworth et~al.(1997)Wentworth, Graham, Pike, Beukelman, Ramsey and Barron}]{landslide}
\bibinfo{author}{Wentworth, C.M.}, \bibinfo{author}{Graham, S.E.}, \bibinfo{author}{Pike, R.J.}, \bibinfo{author}{Beukelman, G.S.}, \bibinfo{author}{Ramsey, D.W.}, \bibinfo{author}{Barron, A.D.}, \bibinfo{year}{1997}.
\newblock \bibinfo{title}{Summary distribution of slides and earth flows in the San Francisco Bay Region, California}.
\newblock \bibinfo{type}{Technical Report}. US Dept. of the Interior, US Geological Survey,.
%Type = Article
\bibitem[{Wu et~al.(2012)Wu, Lindell and Prater}]{wu2012logistics}
\bibinfo{author}{Wu, H.C.}, \bibinfo{author}{Lindell, M.K.}, \bibinfo{author}{Prater, C.S.}, \bibinfo{year}{2012}.
\newblock \bibinfo{title}{Logistics of hurricane evacuation in hurricanes katrina and rita}.
\newblock \bibinfo{journal}{Transportation research part F: traffic psychology and behaviour} \bibinfo{volume}{15}, \bibinfo{pages}{445--461}.
%Type = Inproceedings
\bibitem[{Yusoff et~al.(2008)Yusoff, Ariffin and Mohamed}]{yusoff2008optimization}
\bibinfo{author}{Yusoff, M.}, \bibinfo{author}{Ariffin, J.}, \bibinfo{author}{Mohamed, A.}, \bibinfo{year}{2008}.
\newblock \bibinfo{title}{Optimization approaches for macroscopic emergency evacuation planning: a survey}, in: \bibinfo{booktitle}{2008 International symposium on information technology}, \bibinfo{organization}{IEEE}. pp. \bibinfo{pages}{1--7}.

\end{thebibliography}



\end{document}